\documentclass[a4paper]{article}
\usepackage[utf8]{inputenc}
\usepackage{geometry}
\usepackage{microtype}
\usepackage{graphicx}
\usepackage{subfigure}
\usepackage{booktabs} 
\usepackage{microtype}
\usepackage{framed}
\usepackage{graphicx,stfloats}
\usepackage{subfigure}
\usepackage{booktabs} 
\usepackage{amsmath,amssymb,amsfonts}
\allowdisplaybreaks
\usepackage{mathtools}
\usepackage{amsthm}
\allowdisplaybreaks[4]
\usepackage{float,bbm}
\usepackage{titling}
\usepackage{setspace}
\usepackage{algorithm,algorithmic}


\usepackage{hyperref}
\usepackage[capitalize,noabbrev]{cleveref}

\theoremstyle{theorem}
\newtheorem{theorem}{Theorem}[section]

\newtheorem{lemma}{Lemma}[section]

\theoremstyle{definition}
\newtheorem{definition}[theorem]{Definition}

\theoremstyle{remark}
\newtheorem{remark}{Remark}[section]

\usepackage[textsize=tiny]{todonotes}
\hypersetup{hidelinks}

\begin{document}

\begin{center}
	{\Large Convergence Rate Analysis for Optimal Computing Budget Allocation Algorithms}
	\\\vspace{20pt}
	Yanwen Li$^1$~~~~~~~~~~~~~~~~~~~Siyang Gao$^{2,3}$\\
	{~~~~~~~~~~~~\small \url{yanwen.li@hotmail.com}}~~~~{\small \url{siyangao@cityu.edu.hk}}
	\\\hspace{10pt}
	
	\small  
	$^1$ School of Physical and Mathematical Sciences, Nanyang Technological University, Singapore\\
	$^2$ Department of Advanced Design and Systems Engineering, City University of Hong Kong, Hong Kong\\
	$^3$ School of Data Science, City University of Hong Kong, Hong Kong
\end{center}

\vspace{10pt}

\begin{abstract}
	Ordinal optimization (OO) is a widely-studied technique for optimizing discrete-event dynamic systems (DEDS). It evaluates the performance of the system designs in a finite set by sampling and aims to correctly make ordinal comparison of the designs. A well-known method in OO is the optimal computing budget allocation (OCBA). It builds the optimality conditions for the number of samples allocated to each design, and the sample allocation that satisfies the optimality conditions is shown to asymptotically maximize the probability of correct selection for the best design. In this paper, we investigate two popular OCBA algorithms. With known variances for samples of each design, we characterize their convergence rates with respect to different performance measures. We first demonstrate that the two OCBA algorithms achieve the optimal convergence rate under measures of probability of correct selection and expected opportunity cost. It fills the void of convergence analysis for OCBA algorithms. Next, we extend our analysis to the measure of cumulative regret, a main measure studied in the field of machine learning. We show that with minor modification, the two OCBA algorithms can reach the optimal convergence rate under cumulative regret. It indicates the potential of broader use of algorithms designed based on the OCBA optimality conditions.
\end{abstract}

\vspace{5pt}

\section{Introduction}\label{sec1}

Discrete-event dynamic systems (DEDS) widely emerge in the real world, e.g., in manufacturing production \cite{law1997}, transportation \cite{persula1999}, health-care \cite{jun1999}, supply chain operations \cite{terzi2004}, etc. Given the complex mechanism and human-made rules of operation in DEDS, it is difficult to build analytical models for these systems and their performance is often learned by samples, including samples from a simulation model that faithfully describes the operation details of the systems or samples directly taken from experimentation on the real systems. In this research, we suppose there is a finite set of system designs, and we want to correctly select the best design (the design with the maximal mean performance) by sampling. This problem is known as ranking and selection (R\&S), pure exploration or best arm identification under different contexts.

When the mean performance of each design is estimated by sample means, the error diminishes slowly in the order of $1/\sqrt{n}$, where $n$ is the total number of samples. Consequently, it usually requires a large number of samples to achieve a good estimate of the mean performance. On the other hand, samples are costly and the total sampling budget is limited in practice. It is therefore critical to smartly allocate the samples to the designs for correct selection of the best one.

Ordinal optimization (OO) was proposed as an effective technique for this purpose \cite{ho1992,ho2000}. It shifted the research focus from performance estimation of the designs to order comparison. Since OO enjoys an exponential convergence rate for order comparison \cite{dai1996}, a rate significantly faster than that of performance estimation ($1/\sqrt{n}$), it could lead to highly efficient selection of the best design.

With a finite sampling budget, the best design cannot be correctly selected with probability one. To evaluate the evidence of correct selection, two important performance measures were proposed in the literature. The first is the probability of correct selection (PCS). It is defined as the probability that the selected design is identical to the best design. The other is the expected opportunity cost (EOC). It is the expectation of the difference in the mean between the selected design and the real best one. EOC is also called linear loss or simple regret \cite{bubeck2009,audibert2010,chick2010}. This measure considers not only the chance of the correct selection but also the consequence of an incorrect selection, and is usually preferred by risk-neutral practitioners.

PCS and EOC measure the quality of the selection when the sampling budget is used up and are appropriate for samples from simulation models where sampling non-best designs does not cause immediate costs. In contrast to sampling of simulation models, each sample collected from the real system may incur a direct economic cost and the total costs increase as long as the system runs on any non-best design. For such a case, the measure of cumulative regret (CR) is commonly used and is of practical importance in the situations such as sensor management \cite{washburn2008}, clinic trials \cite{ryzhov2012}, ad placement \cite{bubeck2012}, and dynamic pricing \cite{slivkins2019}. CR is defined as the sum of the difference in the mean between the best design and each sampled design as the selection algorithm proceeds. Under CR, the selection algorithms should avoid sampling bad designs too much. Obviously, CR imposes a substantially different criterion for the best design selection when comparing with PFS and EOC. According to \cite{bubeck2009}, the selection algorithms performing well in terms of PCS and EOC may have poor performance in terms of CR.

There are three popular methods for OO. The first is the indifference-zone (IZ) method. IZ is built on the assumption that the difference in the mean between the best and each non-best design is no less than a specified IZ parameter $\delta>0$. It seeks to select the best design with PCS exceeding a certain threshold by the use of as little sampling efforts as possible \cite{kim2001,nelson2001}. The second is the value of information procedure (VIP). It typically adopts a Bayesian framework and allocates samples using predictive distributions of further samples in order to maximize a certain acquisition function such as the expected improvement (EI) or knowledge gradient (KG) \cite{frazier2008,chick2010,ryzhov2016,peng2017}.

The third OO method is the optimal computing budget allocation (OCBA). The goal of this method is to maximize PCS under a sampling budget constraint, where the decision variables are the number of samples allocated to each design. To solve it, Chen et al. \cite{chen2000} replaced PCS by a simple analytical approximation as the objective function. By checking the KKT conditions of the resulting optimization problem, they derived the optimality conditions for sample allocation. If the number of samples allocated to each design follows the optimality conditions, the PCS can be asymptotically maximized. Glynn and Juneja \cite{glynn2004} investigated the same PCS optimization problem using the large deviations theory and achieved the same optimality conditions as in \cite{chen2000}. Gao et al. \cite{gao2017} showed that these optimality conditions asymptotically minimize EOC as well. Following \cite{chen2000} and \cite{glynn2004}, some variants of the R\&S problem have also been studied \cite{chen2008,lee2012,xiao2013,gao2015auto,gao2017auto,xiao2017,xiao2019,pedrielli2019,kou2021,wang2021,li2022,goodwin2022}.           

Although OCBA has various extensions mentioned above and robust empirical performance \cite{chen2000,branke2007,chick2010,chen2011,li2022b}, the OCBA algorithms for R\&S and its variant problems still remain meta-heuristics and lack theoretical analysis and justification. In this paper, we focus on two common OCBA algorithms \cite{chen2011,glynn2004} with Gaussian samples and known variances of each design, and study their convergence rates under different performance measures. The contributions of this paper are summarized as follows:
\begin{itemize}
	\item We analyze the convergence behavior of the two OCBA algorithms with Gaussian samples and known variances of each design. We derive their asymptotic allocation and the convergence rates under PCS and EOC. It fills the gap of theoretical analysis of algorithms designed based on the OCBA optimality conditions and explains the high efficiency of them observed in the numerical testing.
	
	\item We conduct minor modification to the two OCBA algorithms for the measure of CR and show that the modified algorithms achieve the optimal convergence rate under CR. It provides good insights into the connection and difference between PCS, EOC and CR, and implies application potential of OCBA-like algorithms in other types of selection problems.
\end{itemize}

The rest of the paper is organized as follows.  Section \ref{sec2} introduces the two OCBA algorithms under study and the measures of PCS, EOC and CR. Section \ref{sec3} provides the rate analysis of the OCBA algorithms. Section \ref{sec4} proposes modification of the OCBA algorithms for CR and characterizes their convergence rates. Section \ref{sec5} presents the numerical results. Section \ref{sec6} concludes the paper.

\section{Problem Statement}\label{sec2}

In this section, we first review the development of the OCBA optimality conditions in \cite{chen2011,glynn2004}. Then, we describe two OCBA algorithms based on the optimality conditions. Last, we introduce the measures PCS, EOC and CR.

\subsection{Optimality Conditions}

We introduce the following notations.

\renewcommand{\arraystretch}{1}
\begin{tabular}{ll}
	$k$ & the total number of designs;\\
	$n$ & the total number of samples (budget);\\
	$X_{i,l}$ & value of the $l$-th sample for design $i$, $l\in\mathbb{N}$;\\
	$\mu_i$ & mean of $X_{i,l}$, i.e., $\mu_i=\mathbb{E}\left[X_{i,l}\right]$;\\
	$\sigma_i^{2}$ & variance of $X_{i,l}$, i.e., $\sigma_i^{2}={\rm Var}\left[X_{i,l}\right]$;\\
	$b$ & real best design, i.e., $b=\arg\max_{i\in\left\{1,\dots,k\right\}}\mu_i$,\\
	& $\mu_b>\mu_i$ for $\forall i\ne b$;\\
	$\alpha_i$ & proportion of samples allocated to design $i$;\\
	$n_i$ & number of samples allocated to design $i$, i.e., \\
	& $n_i=\alpha_i n$;\\
	$\hat{\mu}_i$ & sample mean of $X_{i,l}$, i.e., $\hat{\mu}_i=\frac{1}{n_i}\sum_{l=1}^{n_i}X_{i,l}$;\\
	$\hat{b}$ & estimated best design, i.e.,  $\hat{b}=\arg\max_{i}\hat{\mu}_i$.
\end{tabular}

The goal is to select the real best design $b$ from the $k$ alternative designs $\left\{1,\dots,k\right\}$ with unknown means.
Sample $X_{i,l}$'s follow Gaussian distribution $\mathcal{N}\left(\mu_i,\sigma_i^{2}\right)$ with known variance $\sigma_i^{2}$, $i\in\left\{1,\dots,k\right\}$, $l\in\mathbb{N}$. We assume that the samples are identically and independently distributed from replication to replication for the same design and independent across different designs. Given the fixed budget $n$, we sequentially sample each design and output the estimated best design $\hat{b}\in\left\{1,\dots,k\right\}$ when the sampling budget is used up.

The probability of correct selection (PCS) is the probability that the selected design $\hat{b}$ equals the real best design $b$. Note that PCS is a function of sample allocation $n_i$, $i=1,\dots,k$. To determine the optimal sample allocation, we have the following optimization problem,
\begin{equation}\label{opt-pcs}
	\begin{aligned}
		\max_{n_1,\dots,n_k} \text{PCS}~~
		\text{s.t.}~\sum_{i=1}^{k}n_i=n,n_i\geq 0~{\rm for}~\forall i.
	\end{aligned}
\end{equation}
PCS does not have an analytical expression. To overcome the difficulty, Chen et al. \cite{chen2000} replaced PCS with a simple analytical approximation. By utilizing the Karush-Kuhn-Tucker (KKT) conditions and assuming $n_b\gg n_i$ for $\forall i\ne b$, they derived the following equations for sample allocation,
\begin{align}\label{ocba-chen}
	\frac{n_{b}^{2}}{\sigma_{b}^{2}}-\sum_{i\ne b}\frac{n_i^{2}}{\sigma_{i}^{2}}=0,~~
	\frac{n_i}{n_j}=\frac{\sigma_i^{2}\left(\mu_b-\mu_j\right)^{2}}{\sigma_j^{2}\left(\mu_b-\mu_i\right)^{2}},~ i,j\ne b.
\end{align}
In addition, Glynn and Juneja \cite{glynn2004} applied the large deviations (LD) theory to study the optimization problem (\ref{opt-pcs}). Denote by $\Lambda_i\left(\theta\right)=\log\left(\mathbb{E}\left[e^{\theta X_{i,l}}\right]\right)$ the log-moment generating function of $X_{i,l}$,  by $J_i\left(\gamma\right)=\sup_{\theta\in\mathbb{R}}\left(\theta\gamma-\Lambda_i\left(\theta\right)\right)$ the Fenchel-Legendre transform of $\Lambda_i\left(\theta\right)$. Glynn and Juneja \cite{glynn2004} considered the convergence rate of the probability of false selection (PFS, equals $1-$PCS) and showed that $\lim_{n\to\infty}\frac{1}{n}\log\left({\rm PFS}\right)=-\min_{i\ne b}R_{i,b}$, where $R_{i,b}=\inf_{\gamma}\left(\alpha_i J_i\left(\gamma\right)+\alpha_b J_b\left(\gamma\right)\right)$, $i\ne b$. Then, the optimization problem (\ref{opt-pcs}) becomes
\begin{align}\label{opt-pfs}
	\max_{\alpha_1,\dots,\alpha_k}\left[\min_{i\ne b}R_{i,b}\right]~
	\text{s.t.}~\sum_{i=1}^{k}\alpha_i=1,\alpha_i\geq 0~{\rm for}~\forall i.
\end{align}
By assuming Gaussian samples $X_{i,l}\sim\mathcal{N}\left(\mu_i,\sigma_i^{2}\right)$, $i=1,\dots,k$, $l\in\mathbb{N}$ and applying the KKT conditions, the following equations for sample allocation $\alpha_i$'s can be obtained:

\begin{align}\label{ocba-glynn}
	\frac{\alpha_{b}^{2}}{\sigma_{b}^{2}}-\sum_{i\ne b}\frac{\alpha_i^{2}}{\sigma_i^{2}}=0,~
	\frac{\left(\mu_{b}-\mu_i\right)^{2}}{\frac{\sigma_i^{2}}{\alpha_i}+\frac{\sigma_{b}^{2}}{\alpha_{b}}}
	=\frac{\left(\mu_{b}-\mu_j\right)^{2}}{\frac{\sigma_j^{2}}{\alpha_j}+\frac{\sigma_{b}^{2}}{\alpha_{b}}},~i,j\ne b.
\end{align}
Sample allocations satisfying (\ref{ocba-glynn}) asymptotically minimize PFS. Note that equations (\ref{ocba-chen}) and (\ref{ocba-glynn}) only have slight difference. If we assume $n_b\gg n_i$ for $i\ne b$, the two sets of equations become identical. For presentation simplicity, we refer both (\ref{ocba-chen}) and (\ref{ocba-glynn}) as optimality conditions. Gao et al. \cite{gao2017} further investigated the minimization of EOC and showed that the sample allocations satisfying (\ref{ocba-glynn}) asymptotically minimize EOC as well.

\subsection{Algorithm Description}

In this section, we introduce two OCBA algorithms designed based on (\ref{ocba-chen}) and (\ref{ocba-glynn}), called OCBA-1 Algorithm and OCBA-2 Algorithm respectively. To facilitate presentation, we introduce some additional notations.

\begin{tabular}{ll}
	$t$ & iteration index of the algorithm;\\
	$I_t$ & design sampled by the algorithm in iteration $t$;\\
	$\hat{\mu}_i^{(t)}$ & sample mean of design $i$ in iteration $t$;\\
	$\hat{b}^{(t)}$ & design with the largest sample mean in iteration $t$;\\
	$\alpha_i^{(t)}$ & the proportion of samples allocated to design $i$\\
	& until iteration $t$;\\
	$N_i^{(t)}$ & the number of samples so far allocated to design $i$\\
	& until iteration $t$.
\end{tabular}

The sample allocation $\alpha_1^{*},\dots,\alpha_k^{*}$ that solves (\ref{ocba-chen}) has an analytical form $\alpha_i^{*}=\frac{\beta_i}{\sum_{j=1}^{k}\beta_j}$ for $i=1,\dots,k$, where $\beta_b=\sigma_b\sqrt{\sum_{j\ne b}\frac{\sigma_j^{2}}{\left(\mu_b-\mu_j\right)^{4}}}$, $\beta_j=\frac{\sigma_j^{2}}{\left(\mu_b-\mu_j\right)^{2}}$ for $j\ne b$. In the selection algorithm, $\alpha_i^{*}$ is unknown and is replaced by its estimate, that is, $\hat{\alpha}_{i}^{(t)}=\frac{\hat{\beta}_{i}^{(t)}}{\sum_{j=1}^{k}\hat{\beta}_{j}^{(t)}}$, where $\hat{\beta}_{\hat{b}^{(t)}}^{(t)}=\sigma_{\hat{b}^{(t)}}\sqrt{\sum_{i\ne\hat{b}^{(t)}}\frac{\sigma_i^{2}}{\left(\hat{\mu}_{\hat{b}^{(t)}}^{(t)}-\hat{\mu}_i^{(t)}\right)^{4}}}$, $\hat{\beta}_{j}^{(t)}=\frac{\sigma_j^{2}}{\left(\hat{\mu}_{\hat{b}^{(t)}}^{(t)}-\hat{\mu}_j^{(t)}\right)^{2}}$ for $j\ne\hat{b}^{(t)}$. Chen and Lee \cite{chen2011} proposed the OCBA-1 Algorithm based on (\ref{ocba-chen}). We show a generalized version of it by introducing a one-time computing budget increment $\Delta\geq 1$.

\begin{framed}
	\textbf{OCBA-1 Algorithm}
	\begin{itemize}
		\item[1:] Input $k$, $n$, $\Delta$, $n_0$.
		\item[2:] Collect $n_0$ samples for each design.
		\item[3:] $t\leftarrow 0$, $N_1^{(t)}=\cdots=N_k^{(t)}=n_0$.
		\item[4:] \textbf{WHILE} $\sum_{i=1}^{k}N_i^{(t)}<n$ \textbf{DO}
		\item[5:] Update $\hat{\mu}_i^{(t)}$ for each $i$, $\hat{b}^{(t)}=\arg\max_i\hat{\mu}_i^{(t)}$.
		\item[6:]  Compute $\hat{\alpha}_i^{(t)}$,  $\hat{N}_i^{(t)}=\hat{\alpha}_i^{(t)}\left(1+\sum_{i=1}^{k}N_i^{(t)}\right)$ for each $i$.
		\item[7:] Find $I_{t}=\arg\max_i\left(\hat{N}_i^{(t)}-N_i^{(t)}\right)$.
		\item[8:] Collect one more sample for design $I_{t}$.
		\item[9:]  $N_{I_{t}}^{(t+1)}=N_{I_{t}}^{(t)}+\Delta$, $N_{i}^{(t+1)}=N_{i}^{(t)}$ for $\forall i\ne I_{t}$, $t\leftarrow t+1$.
		\item[10:] \textbf{END WHILE}
		\item[11:] Output  $\hat{b}^{(n)}=\arg\max_i\hat{\mu}_i^{(n)}$.
	\end{itemize}
\end{framed}


Using (\ref{ocba-glynn}), the OCBA-2 Algorithm was proposed as follows \cite{gao2017}.

\begin{framed}
	\textbf{OCBA-2 Algorithm}
	\begin{itemize}
		\item[1:] Input $k$, $n$, $\Delta$, $n_0$.
		\item[2:] Collect $n_0$ samples for each design.
		\item[3:] $t\leftarrow 0$, $N_1^{(t)}=\cdots=N_k^{(t)}=n_0$.
		\item[4:] \textbf{WHILE} $\sum_{i=1}^{k}N_i^{(t)}<n$ \textbf{DO}
		\item[5:] Update $\hat{\mu}_i^{(t)}$ for each $i$, $\hat{b}^{(t)}=\arg\max_i\hat{\mu}_i^{(t)}$.
		\item[6:] \textbf{IF} $\left(\frac{N_{\hat{b}^{(t)}}^{(t)}}{\sigma_{\hat{b}^{(t)}}}\right)^{2}-\sum_{i\ne \hat{b}^{(t)}}\left(\frac{N_{i}^{(t)}}{\sigma_i}\right)^{2}<0$ \textbf{THEN}
		\item[7:] Set $I_{t}=\hat{b}^{(t)}$.
		\item[8:] \textbf{ELSE}
		\item[9:] Find $I_{t}=\arg\min_{i\ne \hat{b}^{(t)}} \left(\frac{\left(\hat{\mu}_{\hat{b}^{(t)}}^{(t)}-\hat{\mu}_i^{(t)}\right)^{2}}{\frac{\sigma_i^{2}}{N_i^{(t)}}+\frac{\sigma_{\hat{b}^{(t)}}^{2}}{N_{\hat{b}^{(t)}}^{(t)}}}\right)$.
		\item[10:] \textbf{END IF}
		\item[11:] Collect one more sample for design $I_{t}$.
		\item[12:]  $N_{I_{t}}^{(t+1)}=N_{I_{t}}^{(t)}+\Delta$, $N_{i}^{(t+1)}=N_{i}^{(t)}$ for $\forall i\ne I_{t}$, $t\leftarrow t+1$.
		\item[13:] \textbf{END WHILE}
		\item[14:] Output  $\hat{b}^{(n)}=\arg\max_i\hat{\mu}_i^{(n)}$.
	\end{itemize}
\end{framed}


\begin{remark}
	The OCBA-1 and OCBA-2 Algorithms are very different, in spite that they have a similar structure. First, the allocation rule (\ref{ocba-chen}) of OCBA-1 was derived to maximize a finite-sample lower bound of PCS while the rule (\ref{ocba-glynn}) of OCBA-2 was derived to maximize the asymptotic convergence rate of PFS. Second, the implementation of the rules (\ref{ocba-chen}) and (\ref{ocba-glynn}) are different. To be more specific, (\ref{ocba-chen}) has an explicit solution $\left\{\alpha_{i}^{*}\left|i=1,\dots,k\right.\right\}$ and can guide the allocation by the use of Lines 6-7 in the OCBA-1 Algorithm. In contrast, the solution of (\ref{ocba-glynn}) is implicit, and thus we need to use Lines 6-10 in the OCBA-2 Algorithm to select the sampled designs. Last, the numerical performance of the OCBA-1 Algorithm is different from that of OCBA-2. As will be shown in Section \ref{sec5}, the performance of the OCBA-1 Algorithm under PFS and EOC may be competitive with or better than that of the OCBA-2 Algorithm when given a small number of samples, and then OCBA-2 may perform better than OCBA-1 under PFS and EOC when the budget is large enough.
	\hfill $\square$
\end{remark}

\begin{remark}
	The selection of $\Delta$ may influence the finite-sample performance of the OCBA algorithms. According to \cite{chen2011}, the OCBA algorithms with a large $\Delta$ may not be able to generate a good budget allocation. While the OCBA algorithms with a small $\Delta$ may need to allocate samples based on the allocation rules (\ref{ocba-chen}) and (\ref{ocba-glynn}) frequently, and may spend quite some time in the calculations of (\ref{ocba-chen}) and (\ref{ocba-glynn}). Compared with using a large $\Delta$, an advantage of using a small $\Delta$ is that it may result in the algorithm producing a more efficient allocation of the sampling budget and a more satisfactory performance under PFS and EOC.  A recommended choice of $\Delta$ from \cite{chen2011} is a number smaller than 100 or $10\%$ of $k$, and $\Delta=1$ is a sound choice for solving complicated real-world problems with high costs of sampling.
	\hfill $\square$
\end{remark}

\subsection{Performance Measures}\label{sec2.2}

Three performance measures are commonly used to evaluate the quality of the selection for the best design, PFS, EOC and CR. PFS is the probability that the estimated best design $\hat{b}$ does not equal the real best design $b$. PFS in the $t$-th iteration of the selection algorithm can be expressed as 
\begin{align*}
	{\rm PFS}_{t}=\mathbb{P}\left(\hat{b}^{(t)}\ne b\right).
\end{align*}
EOC is the expectation of the difference in the mean between the real best $b$ and the estimated best design $\hat{b}$. In the $t$-th iteration of the selection algorithm, 
\begin{align*}
	{\rm EOC}_t=\mathbb{E}\left[\mu_b-\mu_{\hat{b}^{(t)}}\right].
\end{align*}
CR is the sum of the difference in the mean between the real best design and the sampled design in each iteration of the algorithm. CR in the $t$-th iteration of the selection algorithm can be expressed as 
\begin{align*}
	{\rm CR}_t=t\mu_b-\sum_{s=1}^{t}\mathbb{E}\left[\mu_{I_s}\right].
\end{align*}

\section{Asymptotic Analysis on the OCBA Algorithms}\label{sec3}

In this section, we first introduce some definitions in asymptotic analysis and a lemma showing a convergence property of real sequences. 
Then, we present the convergence rate results of the OCBA-1 and OCBA-2 Algorithms.
\begin{definition}\label{def}
	For two positive real-value sequences $\left\{a_t\big{|}t\in\mathbb{N}\right\}$ and $\left\{b_t\big{|}t\in\mathbb{N}\right\}$,
	\begin{itemize}
		\item[(i)] $a_t$ and $b_t$ are called logarithmically equivalent if $\lim_{t\to\infty}\frac{1}{t}\log \left(\frac{a_t}{b_t}\right)=0$, denoted by $a_t\doteq b_t$.
		
		\item[(ii)] $a_t$ and $b_t$ are called asymptotically equivalent if $\lim_{t\to\infty}\frac{a_t}{b_t}=1$, denoted by $a_t\cong b_t$.
		
		\item[(iii)] $a_t$ is asymptotically bounded both above and below by $b_t$ if $\mathop{\lim\sup}_{t\to\infty}\frac{a_t}{b_t}<\infty$ and $\mathop{\lim\inf}_{t\to\infty}\frac{a_t}{b_t}>0$, denoted by $a_t=\Theta\left(b_t\right)$.
	\end{itemize}
\end{definition}

\begin{lemma}\label{lem1}
	Let $\left\{N_i^{(t)}\Big{|}i=1,\dots,k,t=1,2,\dots\right\}$ be a sequence of positive integers that satisfies $N_i^{(t)}\to\infty$ as $t\to\infty$. Denote by $\alpha_i^{(t)}=\frac{N_i^{(t)}}{\sum_{j=1}^{k}N_j^{(t)}}$ and $M_i^{(t)}=M_{i}^{(t-1)}+\mathbbm{1}\left\{N_i^{(t)}>N_i^{(t-1)}\right\}$ for $\forall i$, where $\mathbbm{1}\left\{\cdot\right\}$ is an indicator function and $M_i^{(0)}=0$ for $\forall i$. 
	\begin{itemize}
		\item[(i)] If each subsequence $\left\{\alpha_i^{\left(t_p\right)}\Big{|}M_i^{\left(p\right)}\to\infty~\text{as}~p\to\infty,i=\right.$ $1,\dots,k,p=1,2,\dots\Big{\}}$ of $\left\{\alpha_i^{(t)}\Big{|}i=1,\dots,k\right\}$ has a convergent subsequence and the convergent subsequence converges to $\left\{\alpha_i\big{|}i=1,\dots,k\right\}$, then  $\lim_{t\to\infty}\alpha_i^{\left(t\right)}=\alpha_i$ for $\forall i$.
		
		\item[(ii)] If each subsequence $\left\{\alpha_i^{\left(t_p\right)}\Big{|}M_i^{\left(p\right)}\to\infty~\text{as}~p\to\infty,i=\right.$ $1,\dots,k,p=1,2,\dots\Big{\}}$ of $\left\{\alpha_i^{(t)}\Big{|}i=1,\dots,k\right\}$ has a convergent subsequence and the convergence point of the convergent subsequence satisfies the optimality conditions (\ref{ocba-glynn}), then 
		\begin{equation}\label{lem1-eq1}
			\begin{aligned}
				&\lim_{t\to\infty}\left(\frac{\alpha_b^{\left(t\right)}}{\sigma_b}\right)^{2}-\sum_{i\ne b}\left(\frac{\alpha_i^{\left(t\right)}}{\sigma_i}\right)^{2}=0,\\
				&\lim_{t\to\infty}\frac{\left(\mu_b-\mu_i\right)^{2}}{\frac{\sigma_i^{2}}{\alpha_i^{\left(t\right)}}+\frac{\sigma_b^{2}}{\alpha_b^{\left(t\right)}}}-\frac{\left(\mu_b-\mu_j\right)^{2}}{\frac{\sigma_j^{2}}{\alpha_j^{\left(t\right)}}+\frac{\sigma_b^{2}}{\alpha_b^{\left(t\right)}}}=0,~\forall i,j\ne b.
			\end{aligned}
		\end{equation} 
	\end{itemize} 
\end{lemma}

The OCBA-1 and OCBA-2 Algorithms are designed based on optimality conditions (\ref{ocba-chen}) and (\ref{ocba-glynn}) respectively. Although conditions (\ref{ocba-chen}) and (\ref{ocba-glynn}) indicate efficient sample allocations for maximizing PCS and minimizing EOC, they do not guarantee that the resulting algorithms will have good performance. The performance of the selection algorithms also depends on whether the sample allocations generated from the algorithms can converge to the OCBA allocations calculated by the optimality conditions and how fast the former converge to the latter. To this end, the OCBA-1 and OCBA-2 Algorithms are meta-heuristics, and theoretical analysis on their convergence behavior is critical to assess their performance and justify the use of them in practice. Theorems \ref{thm-ocba1-sampalloc}-\ref{thm-ocba2-convrate} below serve this purpose.

\setcounter{theorem}{0}
\begin{theorem}\label{thm-ocba1-sampalloc}
	For the OCBA-1 Algorithm, the following statements hold (``a.s.'' means ``almost surely''):
	\begin{itemize}
		\item[(i)] $\lim_{t\to\infty}\hat{b}^{(t)}\overset{a.s.}{=}b$.
		
		\item[(ii)] $\lim_{t\to\infty}\alpha_i^{(t)}\overset{a.s.}{=}\alpha_i^{*}$, where $\left\{\alpha_i^{*}\Big{|}i=1,\dots,k\right\}$ satisfies optimality conditions (\ref{ocba-chen}).
	\end{itemize}
\end{theorem}

\begin{theorem}\label{thm-ocba2-sampalloc}
	For the OCBA-2 Algorithm, the following statements hold:
	\begin{itemize}
		\item[(i)] $\lim_{t\to\infty}\hat{b}^{(t)}\overset{a.s.}{=}b$.
		
		\item[(ii)] $\lim_{t\to\infty}\alpha_i^{(t)}\overset{a.s.}{=}\alpha_i^{**}$ a.s., where $\left\{\alpha_i^{**}\Big{|}i=1,\dots,k\right\}$ satisfies optimality conditions (\ref{ocba-glynn}).
	\end{itemize}
\end{theorem}

Theorems \ref{thm-ocba1-sampalloc} and \ref{thm-ocba2-sampalloc} show that as the sampling budget $n$ goes to infinity, the estimated best designs of the two algorithms will be identical to the real best designs so that PFS and EOC will converge to 0. In addition, sample allocations of the two algorithms will satisfy optimality conditions (\ref{ocba-chen}) and (\ref{ocba-glynn}). With the convergence result of the two algorithms, Theorems \ref{thm-ocba1-convrate} and \ref{thm-ocba2-convrate} further characterize their convergence rates.

\begin{theorem}\label{thm-ocba1-convrate}
	For the OCBA-1 Algorithm, the following statements hold:
	\begin{itemize}
		\item[(i)] ${\rm PFS}_{t}\doteq e^{-\frac{\eta^{*}}{2}t}$ a.s., where $\eta^{*}=\min_{i\ne b}\frac{\left(\mu_i-\mu_b\right)^{2}\Delta}{\frac{\sigma_i^{2}}{\alpha_i^{*}}+\frac{\sigma_b^{2}}{\alpha_b^{*}}}$.
		
		\item[(ii)] ${\rm EOC}_t\doteq e^{-\frac{\eta^{*}}{2}t}$ a.s.
		
		\item[(iii)] ${\rm CR}_t\cong\sum_{i\ne b}\left(\mu_b-\mu_i\right)\alpha_i^{*}\Delta\cdot t$ a.s.
	\end{itemize}
\end{theorem}

\begin{theorem}\label{thm-ocba2-convrate}
	For the OCBA-2 Algorithm, the following statements hold:
	\begin{itemize}
		\item[(i)] ${\rm PFS}_{t}\doteq e^{-\frac{\eta^{**}}{2}t}$ a.s., where $\eta^{**}=\min_{i\ne b}\frac{\left(\mu_i-\mu_b\right)^{2}\Delta}{\frac{\sigma_i^{2}}{\alpha_i^{**}}+\frac{\sigma_b^{2}}{\alpha_b^{**}}}$.
		
		\item[(ii)] ${\rm EOC}_t\doteq e^{-\frac{\eta^{**}}{2}t}$ a.s.
		
		\item[(iii)] ${\rm CR}_t\cong\sum_{i\ne b}\left(\mu_b-\mu_i\right)\alpha_i^{**}\Delta\cdot t$ a.s.
	\end{itemize}
\end{theorem}

According to Definitions 3 and 6 in \cite{boyd2000} and the results of Theorems \ref{thm-ocba1-convrate} and \ref{thm-ocba2-convrate}, PFS and EOC of the OCBA-1 Algorithm have exponential convergence with the same asymptotic rate of geometric convergence $\frac{\eta^{*}}{2}$ and those of the OCBA-2 Algorithm have exponential convergence with the same asymptotic rate of geometric convergence $\frac{\eta^{**}}{2}$, where $\eta^{*}$ and $\eta^{**}$ are calculated using the solutions of optimality conditions (\ref{ocba-chen}) and (\ref{ocba-glynn}) respectively. According to \cite{glynn2004} and \cite{gao2017}, $\frac{\eta^{**}}{2}$ is theoretically the optimal convergence rate for PFS and EOC, and the OCBA-2 Algorithm achieves it. $\frac{\eta^{*}}{2}$ corresponds to a rate slightly slower than $\frac{\eta^{**}}{2}$. It suggests that the OCBA-1 Algorithm will perform slightly worse than the OCBA-2 Algorithm when the sampling budget $n$ is large. Theorems \ref{thm-ocba1-convrate} and \ref{thm-ocba2-convrate} also show that CR of the two algorithms increases linearly with iteration index $t$, which is much slower than the optimal logarithmic increasing rate of CR \cite{agrawal1995,burnetas1997}. In particular, with linear increasing, CR per sample will not diminish, but remain constant as the sampling budget $n$ goes to infinity. This result is not surprising because CR is a measure significantly different from PFS and EOC. CR is associated with the consequence when each sample is collected, and calls for a different sample allocation for it to be minimized. In addition, this result aligns with the finding in \cite{bubeck2009} that for any algorithm with ${\rm CR}_t$ bounded by $\Theta\left(t\right)$, its EOC (i.e., simple regret) can achieve an exponential convergence rate at best.

\begin{remark}
	The OCBA-1 and OCBA-2 Algorithms are presented with known variances $\sigma_i^{2}$'s for each design. If $\sigma_i^{2}$'s are unknown, they are typically estimated by $\hat{\sigma}_i^{2}=\frac{1}{n_i-1}\sum_{l=1}^{n_i}\left(X_{i,l}-\hat{\mu}_i\right)^{2}$. In this case, PFS and EOC of OCBA-1 and OCBA-2 may decrease at a polynomial rate instead of an exponential rate. The reason is that the bias between $\hat{\sigma}_i^{2}$'s and $\sigma_i^{2}$'s may increase the likelihood of the false selection (i.e., $\hat{b}\ne b$). Specifically, based on the first equation of (\ref{ocba-chen}) and (\ref{ocba-glynn}), the OCBA algorithms are more likely to sample the estimated best design $\hat{b}$ than the other designs when $\hat{\sigma}_{\hat{b}}\gg\hat{\sigma}_i$ for $\forall i\ne\hat{b}$. According to Lemma 3.3 of \cite{wu2018}, the probability that $\hat{\sigma}_i$ is stuck at an extremely small number for $\forall i\ne\hat{b}$ converges to zero at a polynomial rate. If $\hat{b}\ne b$ and $\hat{\sigma}_{\hat{b}}\gg\hat{\sigma}_i$ for $\forall i\ne\hat{b}$ in some iteration, the false selection may still occur in the next iteration. In such a case, the probability that the false selection keeps happening may converge to zero at a polynomial rate as the iteration number increases.
	\hfill $\square$
\end{remark}

\section{New Algorithms for Cumulative Regret}\label{sec4}

The OCBA-1 and OCBA-2 Algorithms are easy to implement in practice and exhibit high efficiency under PFS and EOC. While according to Theorems \ref{thm-ocba1-convrate}(iii) and \ref{thm-ocba2-convrate}(iii) in Section \ref{sec3} and \cite{bubeck2009}, the OCBA algorithms underperform in minimizing CR. It is reasonable because the OCBA algorithms are designed to minimize PFS or EOC instead of CR. PFS and EOC assess the quality of the final output (the estimated best design) of the algorithms, but CR is concerned about the consequences of decisions (the designs selected for sampling) during the implementation period.

Nevertheless, existing literature on some popular R\&S algorithms has promoted closer ties between R\&S and the measure of CR \cite{frazier2008,ryzhov2012,ryzhov2016,qin2017,li2022a}. In addition, CR has been commonly used in the field of machine learning to evaluate the performance of the selection algorithms for solving complicated real-world problems such as sensor management \cite{washburn2008}, clinic trials \cite{ryzhov2012}, ad placement \cite{bubeck2012}, and dynamic pricing \cite{slivkins2019}. Driven by curiosity, we explore the possibility of modifying the OCBA algorithms to perform well in terms of CR. We introduce the following definition about CR to facilitate presentation.
\begin{definition}\label{def2}
	A selection algorithm is uniformly maximum (UM) convergent if its CR ${\rm CR}_t\cong\sum_{i\ne b}\frac{\mu_b-\mu_i}{{\rm kl}_{i,b}}\log t$, where ${\rm kl}_{i,b}$ is the Kullback-Leibler divergence between the sampling distributions of design $i\ne b$ and design $b$.
\end{definition}


Lai and Robbins \cite{lai1985} derived this logarithmic lower bound of CR (i.e., $\sum_{i\ne b}\frac{\mu_b-\mu_i}{{\rm kl}_{i,b}}\log t$) and demonstrated that no algorithm could achieve a CR smaller than the lower bound in the asymptotic sense. If our goal is to minimize CR and achieve the UM property, the OCBA-1 and OCBA-2 Algorithms still have upside potential for their performance under CR. In this section, we aim to retain the simple algorithm framework and conduct minor modification to certain steps of the two algorithms for them to achieve good performance under CR.

The modification to be made to the OCBA algorithms is inspired by the Epsilon-Greedy Algorithm \cite{auer2002}, which is a simple and effective algorithm for CR. This algorithm comprises two main parts:
\begin{itemize}
	\item[] \textbf{Part 1 (exploration):} \emph{Uniformly sample a design with exploration probability $\epsilon_t$ in the $t$-th iteration.} The algorithm spends a few samples on estimated non-best designs in case that the current estimated best design is not the real best one.
	
	\item[] \textbf{Part 2 (exploitation):} \emph{Sample the design with the largest sample mean with probability $1-\epsilon_t$ in the $t$-th iteration.} With a small $\epsilon_t$, the algorithm spends most of the samples on the design that appears to be the best for reducing CR. To obtain a meaningful bound for CR, a popular setting is to ensure $\epsilon_t\to 0$ as $t\to\infty$, e.g., $\epsilon_t=t^{-1}$ or $\epsilon_t=t^{-\frac{1}{3}}$.
\end{itemize}

Compared to the two OCBA algorithms, the Epsilon-Greedy Algorithm allocates much more samples to the best design and much less samples to the non-best designs. To modify the OCBA algorithms to be effective for CR, we borrow the exploitation part of the Epsilon-Greedy Algorithm and sample the estimated best design with probability $1-\epsilon_t$. With probability $\epsilon_t$, we keep the sampling strategy in the OCBA algorithms. In particular, we introduce 
\begin{align}\label{epsilon_t}
	\epsilon_t=\min\left\{\frac{h_t}{t},1\right\},
\end{align}
where $h_t=\left(\sum_{i\ne \hat{b}^{(t)}}\frac{\hat{\mu}_{\hat{b}^{(t)}}^{(t)}-\hat{\mu}_i^{(t)}}{\hat{{\rm kl}}_{i,\hat{b}^{(t)}}^{(t)}}\right)\frac{\sum_{i\ne\hat{b}^{(t)}}N_i^{(t)}}{\sum_{i\ne \hat{b}^{(t)}}\left(\hat{\mu}_{\hat{b}^{(t)}}^{(t)}-\hat{\mu}_i^{(t)}\right)N_i^{(t)}}$, $\hat{{\rm kl}}_{i,\hat{b}^{(t)}}^{(t)}=\frac{\left(\hat{\mu}_i^{(t)}-\hat{\mu}_{\hat{b}^{(t)}}^{(t)}\right)^{2}+\sigma_i^{2}}{2\sigma_{\hat{b}^{(t)}}^{2}}+\log\left(\frac{\sigma_{\hat{b}^{(t)}}}{\sigma_i}\right)-\frac{1}{2},~\forall i\ne \hat{b}^{(t)}$. In addition, we set $\Delta=1$ in the modified OCBA-1 and modified OCBA-2 algorithms to avoid wasting sampling efforts as much as possible. The modified OCBA algorithms are called OCBA-1-UM Algorithm and OCBA-2-UM Algorithm and are summarized as follows.

\begin{framed}
	\textbf{OCBA-1-UM Algorithm}
	\begin{itemize}
		\item[1:] Input $k$, $n$, $n_0$.
		\item[2:] Collect $n_0$ samples for each design.
		\item[3:] $t\leftarrow 0$, $N_1^{(t)}=\cdots=N_k^{(t)}=n_0$.
		\item[4:] \textbf{WHILE} $\sum_{i=1}^{k}N_i^{(t)}<n$ \textbf{DO}
		\item[5:] Update $\hat{\mu}_i^{(t)}$ for each $i$, $\hat{b}^{(t)}=\arg\max_i\hat{\mu}_i^{(t)}$.
		\item[6:] Uniformly select $u\in[0,1]$.
		\item[7:] \textbf{IF} $u\leq\frac{h_t}{t}$ \textbf{THEN}
		\item[8:]  Compute $\hat{\alpha}_i^{(t)}$, $\hat{N}_i^{(t)}=\hat{\alpha}_i^{(t)}\left(1+\sum_{i=1}^{k}N_i^{(t)}\right)$ for each $i$.
		\item[9:] Find $I_{t}=\arg\max_i\left(\hat{N}_i^{(t)}-N_i^{(t)}\right)$.
		\item[10:] \textbf{ELSE}
		\item[11:] Set $I_{t}=\hat{b}^{(t)}$.
		\item[12:] \textbf{END IF}
		\item[13:] Collect one more sample for design $I_{t}$.
		\item[14:]  $N_{I_{t}}^{(t+1)}=N_{I_{t}}^{(t)}+1$, $N_{i}^{(t+1)}=N_{i}^{(t)}$ for $\forall i\ne I_{t}$, $t\leftarrow t+1$.
		\item[15:] \textbf{END WHILE}
		\item[16:] Output  $\hat{b}^{(n)}=\arg\max_i\hat{\mu}_i^{(n)}$.
	\end{itemize}
\end{framed}

\begin{framed}
	\textbf{OCBA-2-UM Algorithm}
	\begin{itemize}
		\item[1:] Input $k$, $n$, $n_0$.
		\item[2:] Collect $n_0$ samples for each design.
		\item[3:] $t\leftarrow 0$, $N_1^{(t)}=\cdots=N_k^{(t)}=n_0$.
		\item[4:] \textbf{WHILE} $\sum_{i=1}^{k}N_i^{(t)}<n$ \textbf{DO}
		\item[5:] Update $\hat{\mu}_i^{(t)}$ for each $i$, $\hat{b}^{(t)}=\arg\max_i\hat{\mu}_i^{(t)}$.
		\item[6:] Uniformly select $u\in[0,1]$.
		\item[7:] \textbf{IF} $u\leq\frac{h_t}{t}$ \textbf{THEN}
		\item[8:] ~~~~\textbf{IF} $\left(\frac{N_{\hat{b}^{(t)}}^{(t)}}{\sigma_{\hat{b}^{(t)}}}\right)^{2}-\underset{i\ne \hat{b}^{(t)}}{\sum}\left(\frac{N_{i}^{(t)}}{\sigma_i}\right)^{2}<0$ \textbf{THEN}
		\item[9:] ~~~Set $I_{t}=\hat{b}^{(t)}$.
		\item[10:] ~~~\textbf{ELSE}
		\item[11:] ~~~Find $I_{t}=\arg\min_{i\ne \hat{b}^{(t)}} \left(\frac{\left(\hat{\mu}_{\hat{b}^{(t)}}^{(t)}-\hat{\mu}_i^{(t)}\right)^{2}}{\frac{\sigma_i^{2}}{N_i^{(t)}}+\frac{\sigma_{\hat{b}^{(t)}}^{2}}{N_{\hat{b}^{(t)}}^{(t)}}}\right)$.
		\item[12:] ~~~\textbf{END IF}
		\item[13:] \textbf{ELSE}
		\item[14:] Set $I_{t}=\hat{b}^{(t)}$.
		\item[15:] \textbf{END IF}
		\item[16:] Collect one more sample for design $I_{t}$.
		\item[17:]  $N_{I_{t}}^{(t+1)}=N_{I_{t}}^{(t)}+1$, $N_{i}^{(t+1)}=N_{i}^{(t)}$ for $\forall i\ne I_{t}$, $t\leftarrow t+1$.
		\item[18:] \textbf{END WHILE}
		\item[19:] Output  $\hat{b}^{(n)}=\arg\max_i\hat{\mu}_i^{(n)}$.
	\end{itemize}
\end{framed}
\begin{remark}
	We provide some intuition about CR of the OCBA-1-UM and OCBA-2-UM Algorithms increasing at a logarithmic rate. Note that CR can be written as ${\rm CR}_t=\sum_{i\ne b}\left(\mu_b-\mu_i\right)\mathbb{E}\left[N_i^{(t)}\right]$, where $\mathbb{E}\left[N_i^{(t)}\right]$ is the expected number of samples allocated to design $i$ until the $t$-th iteration. Thus, to evaluate the magnitude of CR, we can analyze  $\mathbb{E}\left[N_i^{(t)}\right]$ achieved by the selection algorithms. By setting the exploration probability $\epsilon_t=\Theta\left(\frac{1}{t}\right)$, numbers of samples allocated to the non-best designs in the two algorithms only have logarithmic growth rates with respect to $t$, which leads to ${\rm CR}_t=\Theta\left(\log t\right)$. \hfill $\square$
\end{remark}

To show the performance of the modified OCBA algorithms, we start with the lemma below.
\begin{lemma}\label{lem3}
	(\cite{cesa2008}) Denote by $\left\{L_s\big{|}0\leq L_s\leq 1,s=1,2,\dots\right\}$ a sequence of random variables.
	\begin{itemize}
		\item[(i)] Define the bounded martingale difference sequence $V_s=\mathbb{E}\left[L_s\big{|}L_1,\dots,L_{s-1}\right]-L_s$ and the associated martingale $S_t=V_1+\dots+V_t$ with conditional variance $K_t=\sum_{s=1}^{t}{\rm Var}\left[L_s\big{|}L_1,\dots,L_{s-1}\right]$. Then, for all $r,v\geq 0$, $\mathbb{P}\left(S_t\geq r,K_t\leq v\right)\leq \exp\left\{-\frac{r^{2}}{2v+\frac{2r}{3}}\right\}$.
		
		\item[(ii)] Define another bounded martingale difference sequence $V'_s=-V_s$ and the associated martingale $S'_t=V'_1+\dots+V'_t$ with conditional variance $K_t=\sum_{s=1}^{t}{\rm Var}\left[L_s\big{|}L_1,\dots,L_{s-1}\right]$. Then, for all $r',v'\geq 0$, $\mathbb{P}\left(S'_t\geq r',K_t\leq v'\right)\leq \exp\left\{-\frac{\left(r'\right)^{2}}{2v'+\frac{2r'}{3}}\right\}$.
	\end{itemize}
\end{lemma}

\setcounter{theorem}{0}
Theorem \ref{thm-ocbaum-sampalloc} below shows the consistency and sample allocations of the OCBA-1-UM and OCBA-2-UM Algorithms.
\begin{theorem}\label{thm-ocbaum-sampalloc}
	For the OCBA-1-UM and OCBA-2-UM Algorithms, the following statements hold:
	\begin{itemize}
		\item[(i)] $\lim_{t\to\infty}\hat{b}^{(t)}\overset{a.s.}{=}b$.
		
		\item[(ii)] $\lim_{t\to\infty}\frac{N_b^{(t)}}{t}\overset{a.s.}{=}1$.
		
		\item[(iii)] $\lim_{t\to\infty}\frac{N_i^{(t)}}{h^{*}\log t}\overset{a.s.}{=}\alpha_i^{*}$ for $\forall i\ne b$, where $h^{*}=\frac{\sum_{i\ne b}\frac{\mu_b-\mu_i}{{\rm kl}_{i,b}}}{\sum_{i\ne b}\alpha_i^{*}\left(\mu_b-\mu_i\right)}$, ${\rm kl}_{i,b}=\frac{\left(\mu_i-\mu_b\right)^{2}+\sigma_i^{2}}{2\sigma_b^{2}}+\log\left(\frac{\sigma_b}{\sigma_i}\right)-\frac{1}{2}$ for $\forall i\ne b$ and $\left\{\alpha_i^{*}\Big{|}i=1,\dots,k\right\}$ is the solution of (\ref{ocba-chen}).
	\end{itemize}
\end{theorem}

According to Theorem \ref{thm-ocbaum-sampalloc}, as the budget $n$ goes to infinity, the estimated best designs of the OCBA-1-UM and OCBA-2-UM Algorithms are identical to the real best ones. The real best designs receive almost all the samples and the number of samples allocated to each non-best design has a sublinear growth with the iteration number $t$. In addition, as $n$ goes to infinity, the ratio of the sample sizes for any two non-best designs from the OCBA-1-UM and OCBA-2-UM Algorithms satisfies the solution of the optimality conditions (\ref{ocba-chen}).

\begin{remark}
	We briefly explain why the OCBA-1-UM and OCBA-2-UM Algorithms have the same sample allocations in the limit. The modification made to the OCBA-1 and OCBA-2 Algorithms significantly increases the number of samples allocated to the best design, making it much larger than the number of samples allocated to each non-best design, i.e., $\alpha_{b}\gg\alpha_{i}$ for $\forall i\ne b$. As discussed before, optimality conditions (\ref{ocba-chen}) and (\ref{ocba-glynn}) become identical in this case, so the minor difference in sampling ratios of the non-best designs between (\ref{ocba-chen}) and (\ref{ocba-glynn}) vanishes.
	\hfill $\square$
\end{remark}

\begin{remark}
	Theorem \ref{thm-ocbaum-sampalloc}(ii) means that the best design receives a much greater proportion of the sampling budget than the non-best designs. It suggests that the original OCBA algorithms (i.e., OCBA-1 and OCBA-2) and the modified OCBA algorithms (i.e., OCBA-1-UM and OCBA-2-UM) generally do not have the same sample allocation in the limit. While Theorem \ref{thm-ocbaum-sampalloc}(iii) shows that the modified OCBA algorithms share some similarities with the original OCBA algorithms. That is, for the OCBA-1 Algorithm and the modified OCBA algorithms, the ratios of sample allocations for the non-best designs satisfy that
	\begin{align}\label{rem6-eq1}
		\lim_{t\to\infty}\frac{\alpha_{i,t}}{\alpha_{j,t}}=\frac{\alpha_i^{*}}{\alpha_j^{*}},~\forall i,j\ne b.
	\end{align}	
	Providing that the proportion of the budget allocated to the best design is much larger than that allocated to each non-best design, the sample allocations of the OCBA-2 Algorithm satisfying (\ref{ocba-glynn}) can be very closely approximated by (\ref{rem6-eq1}). In addition, according to \cite{Pasupathy2014}, the allocation (\ref{rem6-eq1}) is close to the optimal allocation when the number of designs becomes large. It implies that the modified OCBA algorithms can be close to optimal when there are a large number of alternative designs.
	\hfill $\square$
\end{remark}

The theorem below further characterizes the convergence rates of the OCBA-1-UM and OCBA-2-UM Algorithms.

\begin{theorem}\label{thm-ocbaum-convrate}
	For the OCBA-1-UM and OCBA-2-UM Algorithms, the following statements hold:
	\begin{itemize}
		\item[(i)] ${\rm PFS}_{t}\doteq t^{-\frac{\rho^{*}}{2}}$ a.s., where $\rho^{*}=\left(\sum_{i=1}^{k}\beta_i\right)^{-1}h^{*}$, $\beta_j=\frac{\sigma_j^{2}}{\left(\mu_b-\mu_j\right)^{2}}$ for $j\ne b$, $\beta_b=\sigma_b\sqrt{\sum_{j\ne b}\frac{\sigma_j^{2}}{\left(\mu_b-\mu_j\right)^{4}}}$.
		
		\item[(ii)] ${\rm EOC}_t\doteq t^{-\frac{\rho^{*}}{2}}$ a.s.
		
		\item[(iii)] ${\rm CR}_t\cong\left(\sum_{i\ne b}\frac{\mu_b-\mu_i}{{\rm kl}_{i,b}}\right)\log t$ a.s.
	\end{itemize}
\end{theorem}

Theorem \ref{thm-ocbaum-convrate} shows that PFS and EOC of the OCBA-1-UM and OCBA-2-UM Algorithms converge to 0 polynomially fast. Note that PFS and EOC of the OCBA-1 and OCBA-2 Algorithms converge to 0 exponentially fast. It implies that the modification of the OCBA-1 and OCBA-2 Algorithms slows down the convergence of the algorithms under PFS and EOC. It makes sense because the OCBA-1-UM and OCBA-2-UM Algorithms are designed for CR and allocate fewer samples to the non-best designs to reduce CR. Such sample allocations will lead to larger PFS and EOC compared to the OCBA-1 and OCBA-2 Algorithms. In addition, according to Definition \ref{def2}, the OCBA-1-UM and OCBA-2-UM Algorithms are UM convergent, i.e., they achieve the optimal convergence rate under CR. Theorem \ref{thm-ocbaum-convrate} is also in line with the result in \cite{bubeck2009} that an algorithm with the logarithmic growth of CR will have PFS and EOC converging to 0 polynomially fast at best.


\section{Numerical Experiments}\label{sec5}

In this section, we conduct numerical experiments on the OCBA algorithms presented in this paper and two algorithms (i.e., Epsilon-Greedy and UCB1-Normal) in \cite{auer2002} designed to minimize CR. In iteration $t=1,2,\dots$, Epsilon-Greedy prescribes to sample with probability $1-\epsilon_t$ the design with the largest sample mean, and with probability $\epsilon_t$ a randomly chosen design, where $\epsilon_t$ is set to (\ref{epsilon_t}) for the numerical experiments. We consider the following two experiments:
\begin{itemize}
	\item Instance 1 (increasing means with increasing variances): $\mu_i=i$ and $\sigma_i=i$ for $i=1,\dots,10$. The best design $b=10$.
	
	\item Instance 2 (increasing means with decreasing variances): $\mu_i=i$ and $\sigma_i=11-i$ for $i=1,\dots,10$. The best design $b=10$.
\end{itemize}
In each experiment, we divide the algorithms to be tested into three groups and compare them numerically within the groups:
\begin{itemize}
	\item Compare the numerical performance of the OCBA \cite{chen2000}, OCBA-1, and OCBA-2 Algorithms for different values of $\Delta$ ($\Delta=1$ and $\Delta=10$). 
	
	\item Compare the numerical performance of OCBA-1 with $\Delta=1$, OCBA-2 with $\Delta=1$, the OCBA-1-UM Algorithm, and the OCBA-2-UM Algorithm.
	
	\item Compare the numerical performance of the OCBA-1-UM Algorithm, the OCBA-2-UM Algorithm, Epsilon-Greedy, and UCB1-Normal.
\end{itemize}
We characterize the sample allocations, PFS, EOC, and CR of the algorithms. To be more specific, we use $\alpha_{i,t}$, $-\frac{1}{t}\log\left({\rm PFS}_t\right)$, $-\frac{1}{t}\log\left({\rm EOC}_t\right)$, $\frac{{\rm CR}_t}{t}$ to describe the sample allocation for design $i$ given the number of samples $t$, and the convergence rates of PFS, EOC, and CR of each algorithm given the number of samples $t$. The initial number of samples allocated to each design is $n_0=5$. We repeat each algorithm five hundred times to obtain its estimated mean performance.

\begin{figure}[!htbp]
	\centering
	\begin{subfigure}
		\centering
		\includegraphics[scale=0.21]{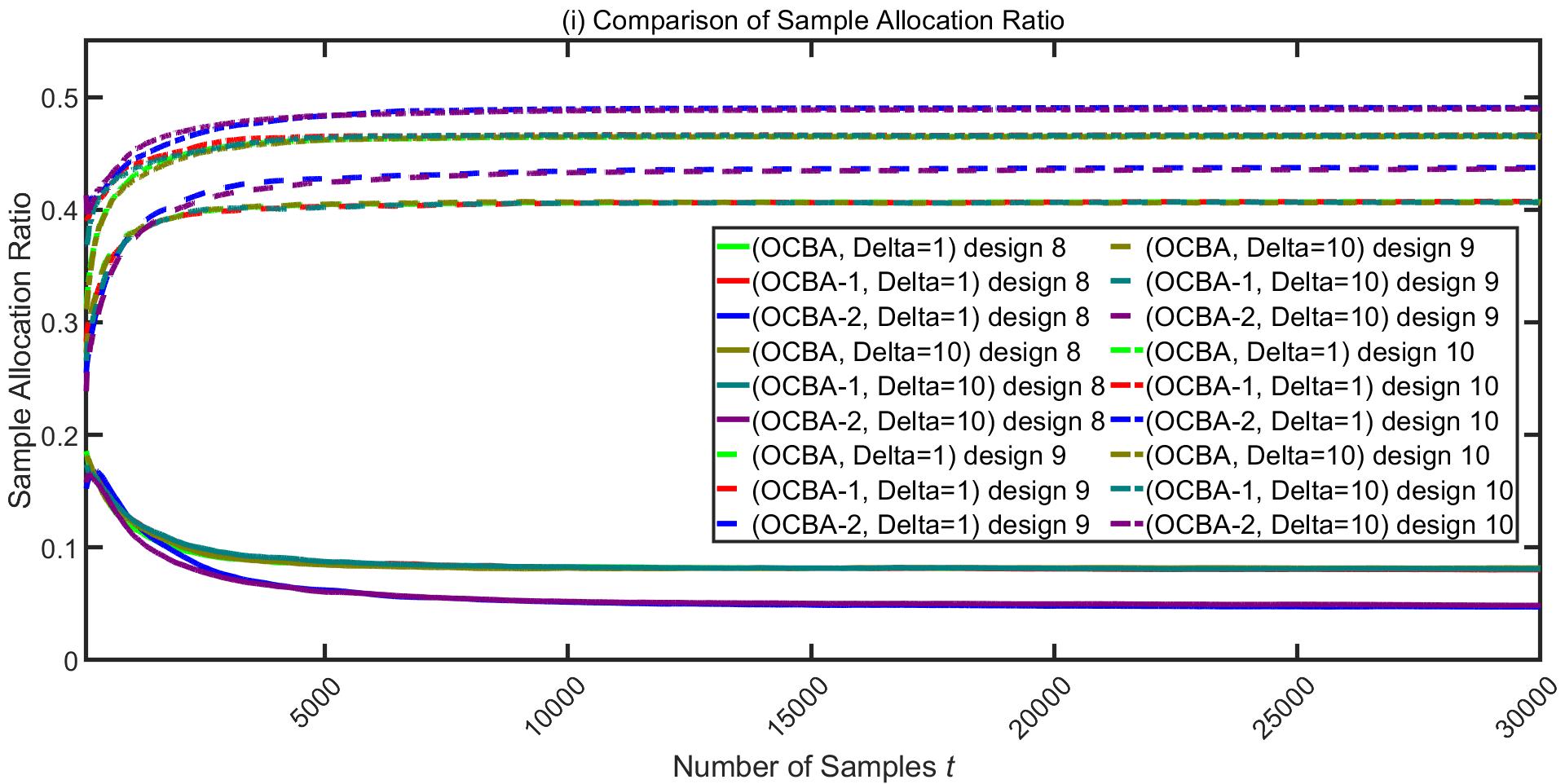}
	\end{subfigure}
	\begin{subfigure}
		\centering
		\includegraphics[scale=0.335]{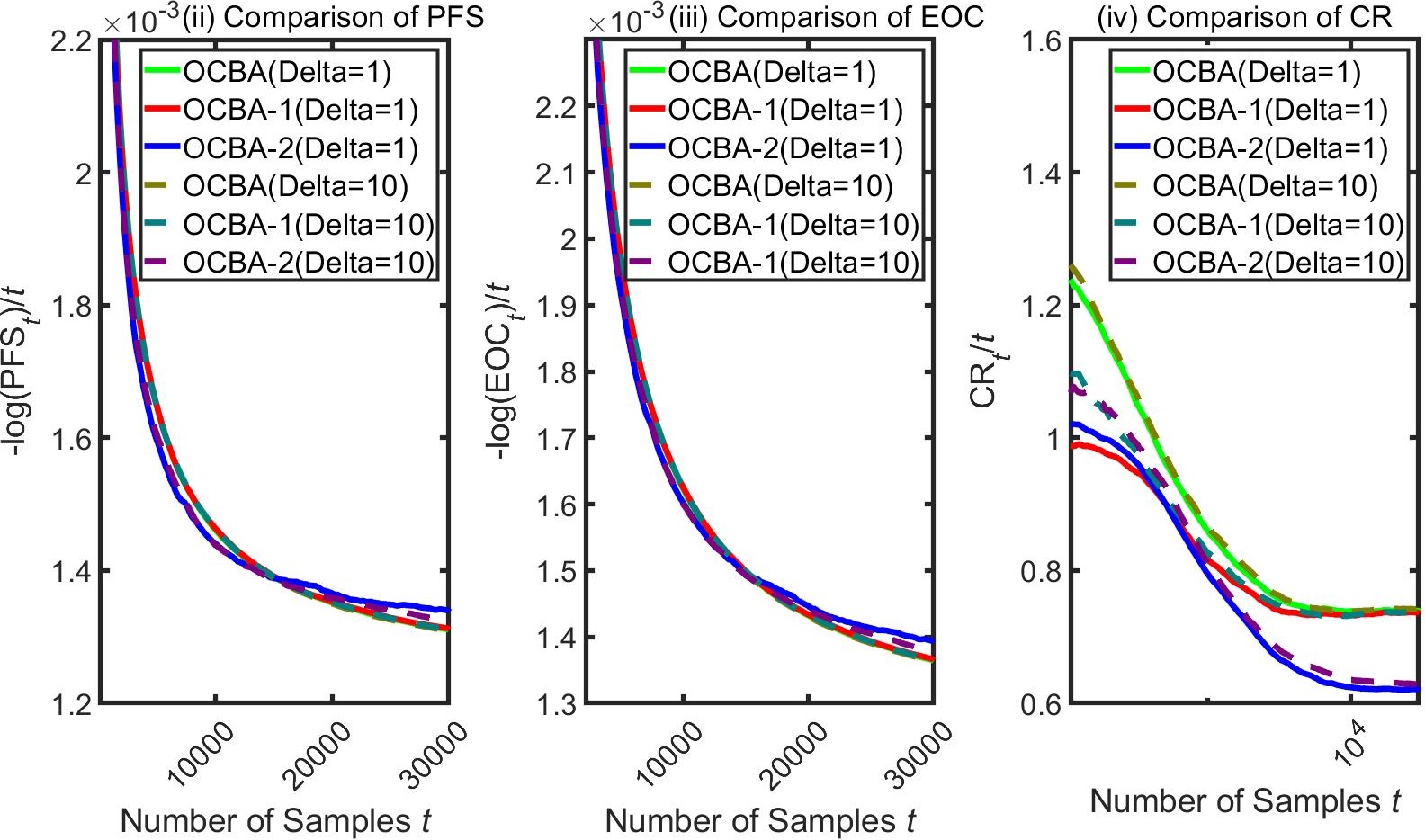}
	\end{subfigure}
	\caption{OCBA algorithms with different $\Delta$'s for instance 1.}
	\label{fig1}
\end{figure}

The sample allocations of OCBA, OCBA-1, and OCBA-2 are shown in Figures \ref{fig1}(i) and \ref{fig3}(i). We select three designs (designs 8, 9, 10) for simplicity of presentation and exhibit their sample allocations. We can see that the sample allocations of OCBA and OCBA-1 are almost the same, no matter whether $\Delta=1$ or $\Delta=10$. OCBA-2 with $\Delta=1$ and OCBA-2 with $\Delta=10$ have almost the same values in terms of sample allocations. However, the sample allocations of OCBA-2 are different from those of OCBA and OCBA-1, no matter whether $\Delta=1$ or $\Delta=10$. 

Figures \ref{fig1}(ii) and \ref{fig1}(iii) show that the values of $-\frac{1}{t}\log\left({\rm PFS}_t\right)$ and $-\frac{1}{t}\log\left({\rm EOC}_t\right)$ for OCBA-2 are slightly smaller than those for OCBA and OCBA-1 when the number of samples is smaller than 15,000. After that, the values of $-\frac{1}{t}\log\left({\rm PFS}_t\right)$ and $-\frac{1}{t}\log\left({\rm EOC}_t\right)$ for OCBA-2 are slightly larger than those for OCBA and OCBA-1. Figures \ref{fig3}(ii) and \ref{fig3}(iii) present that OCBA, OCBA-1 and OCBA-2 have almost the same values in terms of $-\frac{1}{t}\log\left({\rm PFS}_t\right)$ and $-\frac{1}{t}\log\left({\rm EOC}_t\right)$ when the number of samples is smaller than 400. After that, the values of $-\frac{1}{t}\log\left({\rm PFS}_t\right)$ and $-\frac{1}{t}\log\left({\rm EOC}_t\right)$ for OCBA-2 have a clear lead over those for OCBA and OCBA-1. 

In Figure \ref{fig1}(iv), it can be observed that the values of $\frac{{\rm CR}_t}{t}$ for OCBA-2 have been smaller than those for OCBA given any number of samples, and are smaller than those for OCBA-1 when the number of samples is larger than $1,000$. Figure \ref{fig3}(iv) shows that for any number of samples, the values of $\frac{{\rm CR}_t}{t}$ for OCBA-2 have been smaller than those for OCBA and OCBA-1, and OCBA with $\Delta=10$ has the largest values of $\frac{{\rm CR}_t}{t}$. In Figures \ref{fig1}(iv) and \ref{fig3}(iv), the values of $\frac{{\rm CR}_t}{t}$ for OCBA-1 are smaller than those for OCBA given a small number of samples, and then OCBA and OCBA-1 have almost the same values in terms of $\frac{{\rm CR}_t}{t}$. The values of $\frac{{\rm CR}_t}{t}$ for OCBA and OCBA-1 with $\Delta=1$ are slightly smaller than those for OCBA and OCBA-1 with $\Delta=10$, while the gaps among them narrow as the number of samples increases. The similar pattern emerges for OCBA-2 with $\Delta=1$ and $\Delta=10$. 

\begin{figure}[!htbp]
	\vspace{0.5mm}
	\centering
	\begin{subfigure}
		\centering
		\includegraphics[scale=0.41]{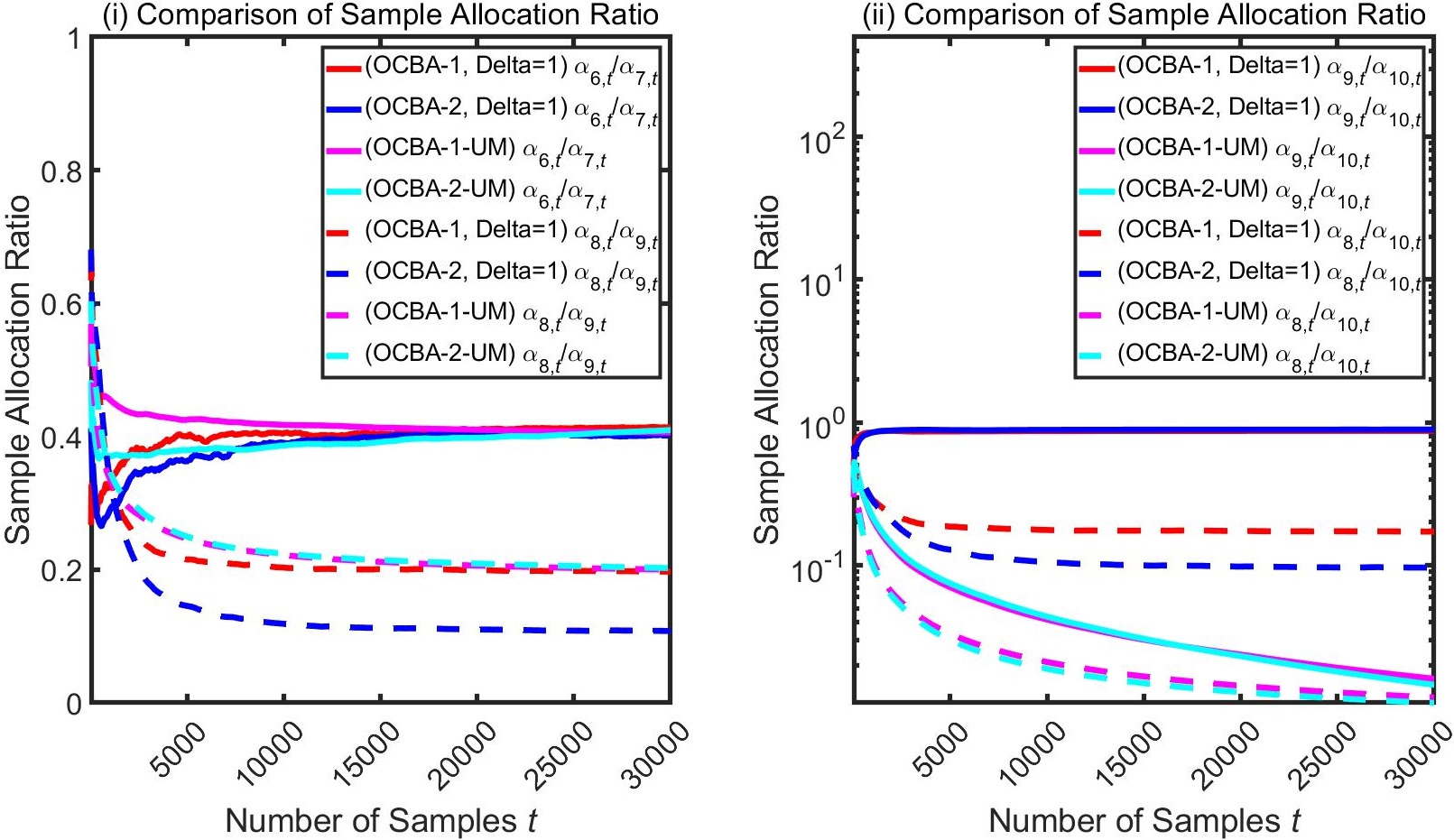}
	\end{subfigure}
	\begin{subfigure}
		\centering
		\includegraphics[scale=0.327]{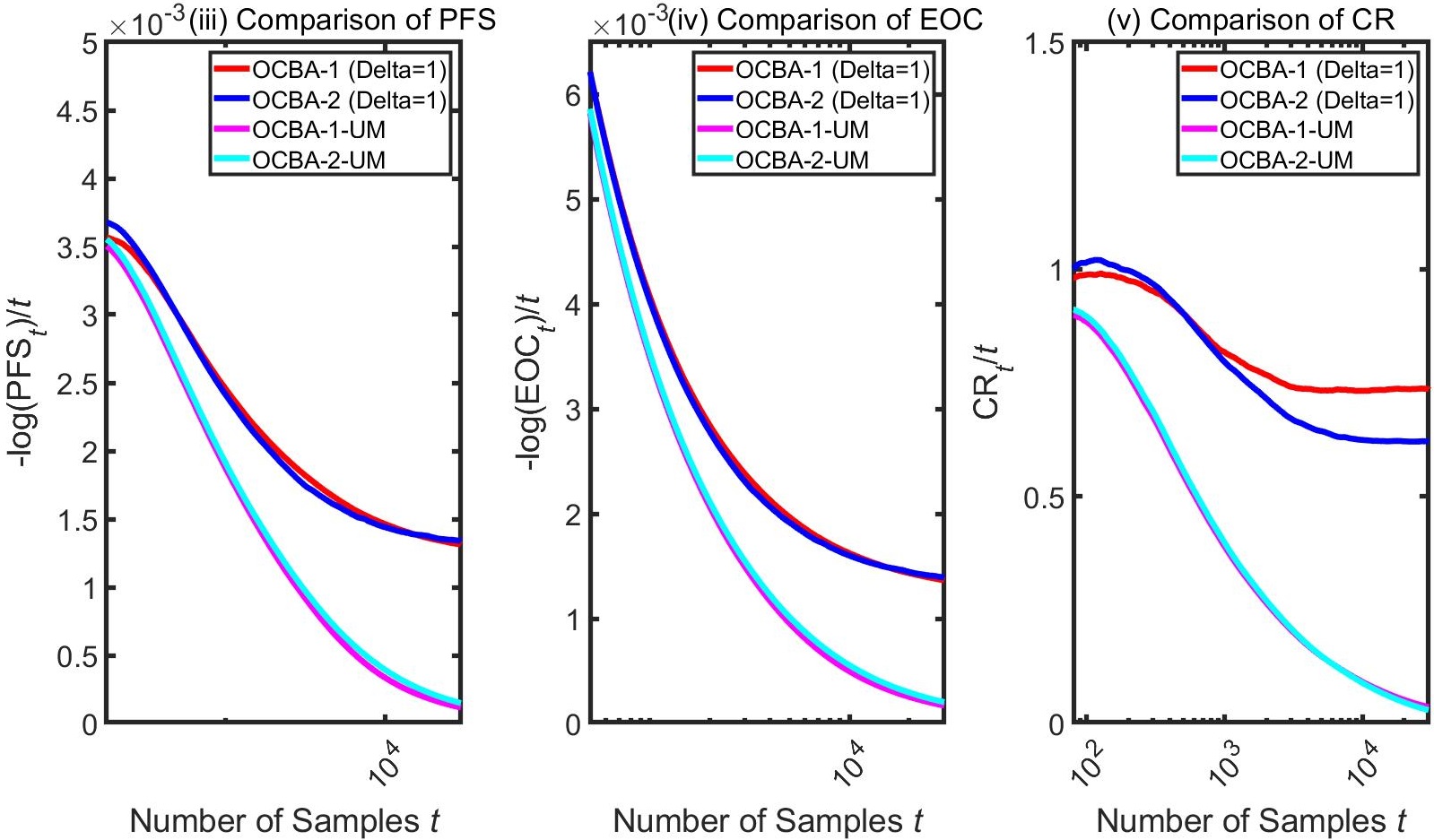}
	\end{subfigure}
	\caption{OCBA-1, OCBA-2, OCBA-1-UM, and OCBA-2-UM for instance 1.}
	\label{fig2}
\end{figure}

Figures \ref{fig2}(i), \ref{fig2}(ii), \ref{fig4}(i), and \ref{fig4}(ii) show the sample allocation behaviors of OCBA-1 with $\Delta=1$, OCBA-2 with $\Delta=1$, OCBA-1-UM, and OCBA-2-UM. For simplicity of presentation, we select some designs (designs 6, 7, 8, 9, 10) and show the ratios of their sample allocations ($\frac{\alpha_{6,t}}{\alpha_{7,t}}$, $\frac{\alpha_{8,t}}{\alpha_{9,t}}$, $\frac{\alpha_{8,t}}{\alpha_{10,t}}$, $\frac{\alpha_{9,t}}{\alpha_{10,t}}$). In Figure \ref{fig2}(i), the gaps among $\frac{\alpha_{6,t}}{\alpha_{7,t}}$'s of OCBA-1, OCBA-1-UM, and OCBA-2-UM vanish as the number of samples is larger than 15,000. Likewise, the gaps among $\frac{\alpha_{8,t}}{\alpha_{9,t}}$'s of OCBA-1, OCBA-1-UM, and OCBA-2-UM vanish as the number of samples is larger than 15,000. The gaps among $\frac{\alpha_{6,t}}{\alpha_{7,t}}$ of OCBA-2 and that of the other OCBA algorithms vanish as the number of samples is larger than 15,000 but there is a clear gap among $\frac{\alpha_{8,t}}{\alpha_{9,t}}$ of OCBA-2 and that of the other OCBA algorithms as the number of samples is larger than 3,000. In Figure \ref{fig2}(ii), $\frac{\alpha_{8,t}}{\alpha_{10,t}}$ and $\frac{\alpha_{9,t}}{\alpha_{10,t}}$ of OCBA-1 and OCBA-2 almost remain constant as the number of samples is larger than 5,000. In contrast, $\frac{\alpha_{8,t}}{\alpha_{10,t}}$ and $\frac{\alpha_{9,t}}{\alpha_{10,t}}$ of OCBA-1-UM and OCBA-2-UM gradually decrease as $t$ grows and are much less than those of OCBA-1 and OCBA-2. The similar pattern emerges in Figures \ref{fig4}(i) and \ref{fig4}(ii).

In Figures \ref{fig2}(iii) and \ref{fig2}(iv), the values of $-\frac{1}{t}\log\left({\rm PFS}_t\right)$ and $-\frac{1}{t}\log\left({\rm EOC}_t\right)$ for OCBA-1-UM and OCBA-2-UM fall faster than those for OCBA-1 and OCBA-2. In addition, $-\frac{1}{t}\log\left({\rm PFS}_t\right)$ and $-\frac{1}{t}\log\left({\rm EOC}_t\right)$ of OCBA-1-UM have almost the same values as those of OCBA-2-UM. The similar pattern emerges in Figures \ref{fig4}(iii) and \ref{fig4}(iv).

Figures \ref{fig2}(v) and \ref{fig4}(v) show that for any number of samples, the values of $\frac{{\rm CR}_t}{t}$ for OCBA-1-UM and OCBA-2-UM have been smaller than those for OCBA-1 and OCBA-2. The gap between $\frac{{\rm CR}_t}{t}$ of OCBA-1 and that of OCBA-1-UM gradually increases as $t$ grows. Likewise, the gap between $\frac{{\rm CR}_t}{t}$ of OCBA-2 and that of OCBA-2-UM gradually increases as $t$ grows. In Figure \ref{fig2}(v), $\frac{{\rm CR}_t}{t}$ of OCBA-1-UM have almost the same values as $\frac{{\rm CR}_t}{t}$ of OCBA-2-UM. In Figure \ref{fig4}(v), the values of $\frac{{\rm CR}_t}{t}$ for OCBA-1-UM are slightly larger than those for OCBA-2-UM when the number of samples is smaller than 100 and then the values of $\frac{{\rm CR}_t}{t}$ for OCBA-1-UM are almost same as those for OCBA-2-UM.

\begin{figure}[!htbp]
	\vspace{0.5mm}
	\centering
	\includegraphics[scale=0.505]{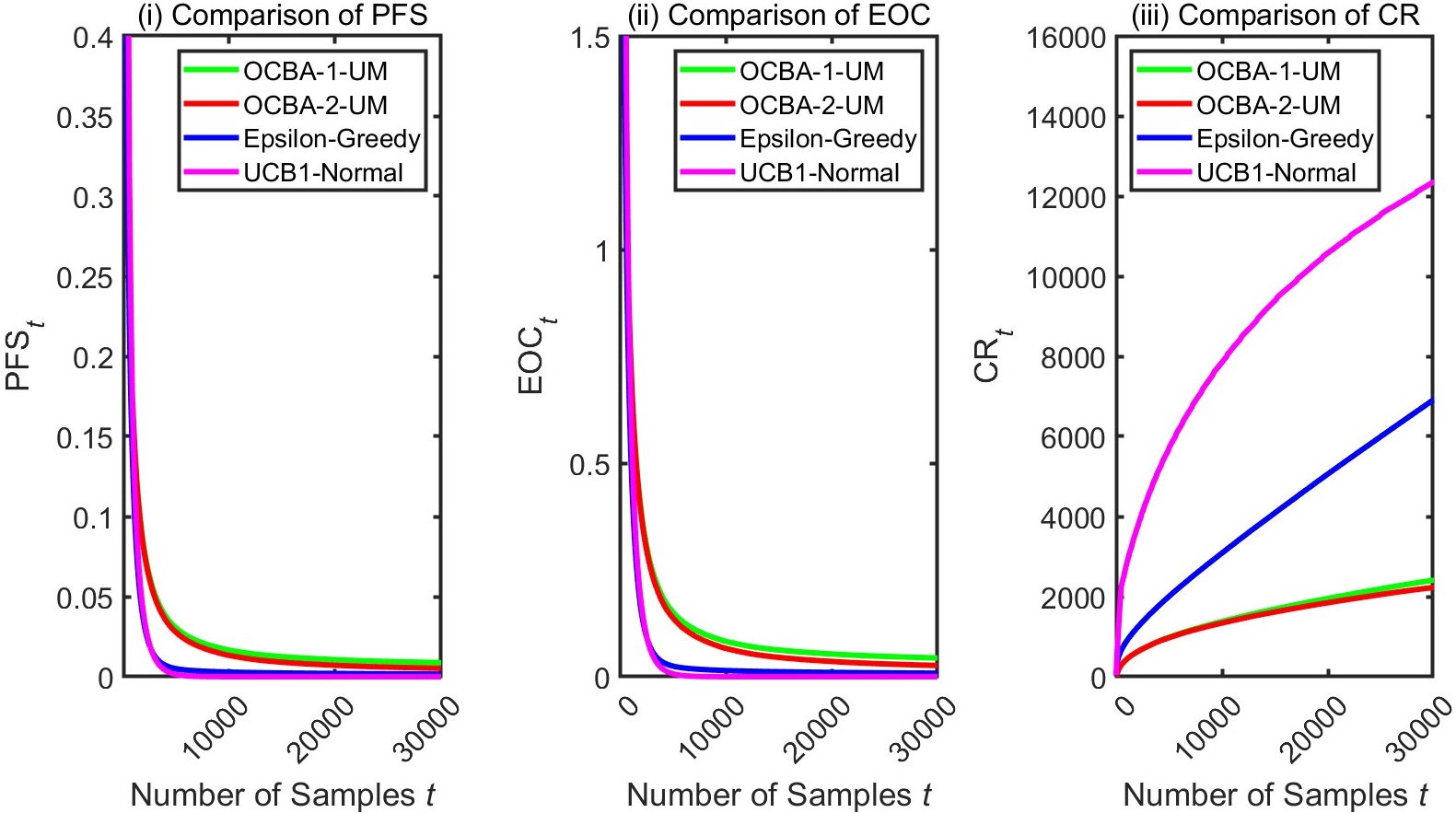}
	\caption{OCBA-1-UM, OCBA-2-UM, Epsilon-Greedy, and UCB1-Normal for instance 1.}
	\label{fig5}
\end{figure}

\begin{figure}[!htbp]
	\centering
	\begin{subfigure}
		\centering
		\includegraphics[scale=0.215]{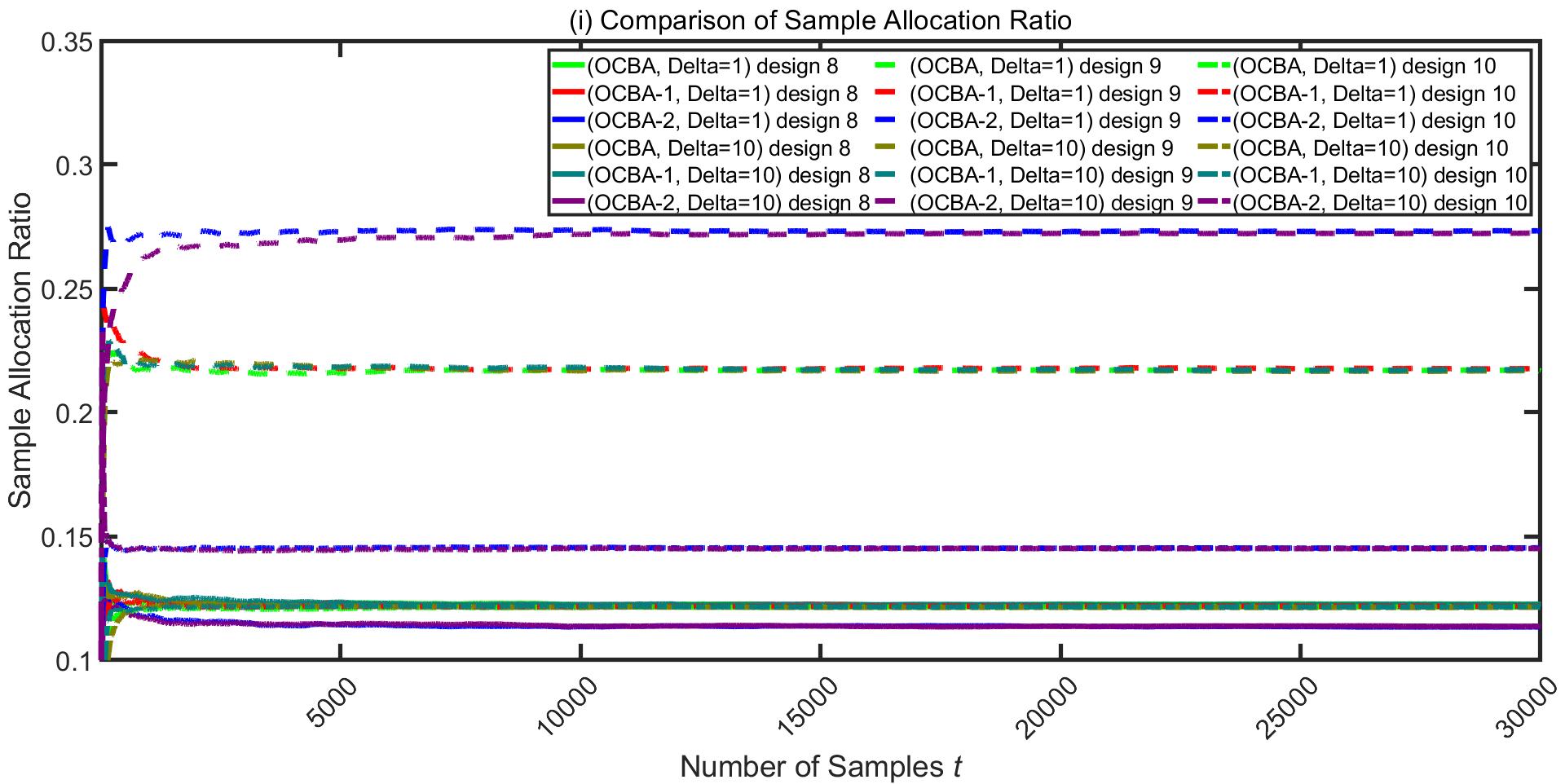}
	\end{subfigure}
	\begin{subfigure}
		\centering
		\includegraphics[scale=0.34]{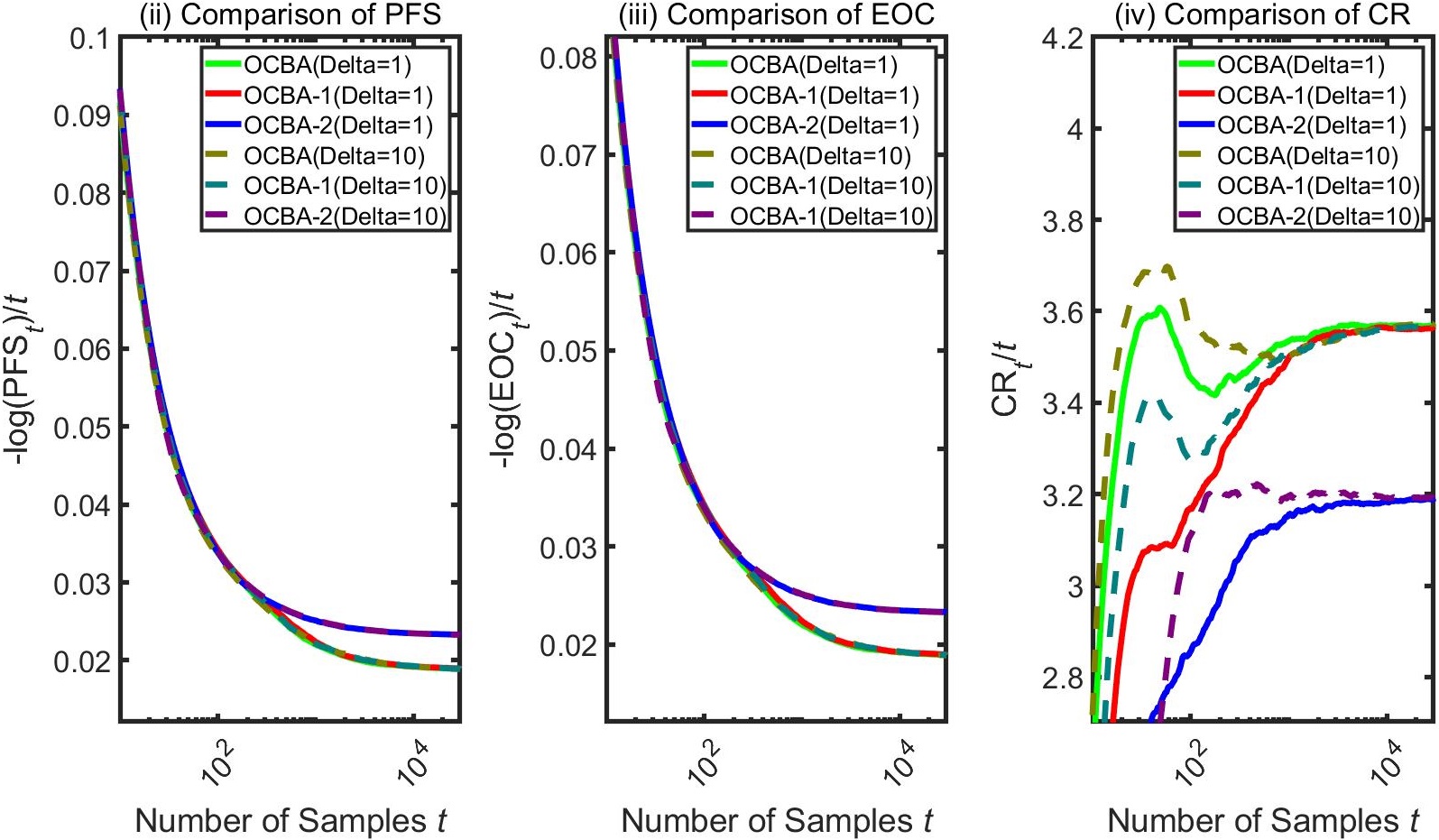}
	\end{subfigure}
	\caption{OCBA algorithms with different $\Delta$'s for instance 2.}
	\label{fig3}
\end{figure}

\begin{figure}[!htbp]
	\vspace{0.5mm}
	\centering
	\begin{subfigure}
		\centering
		\includegraphics[scale=0.42]{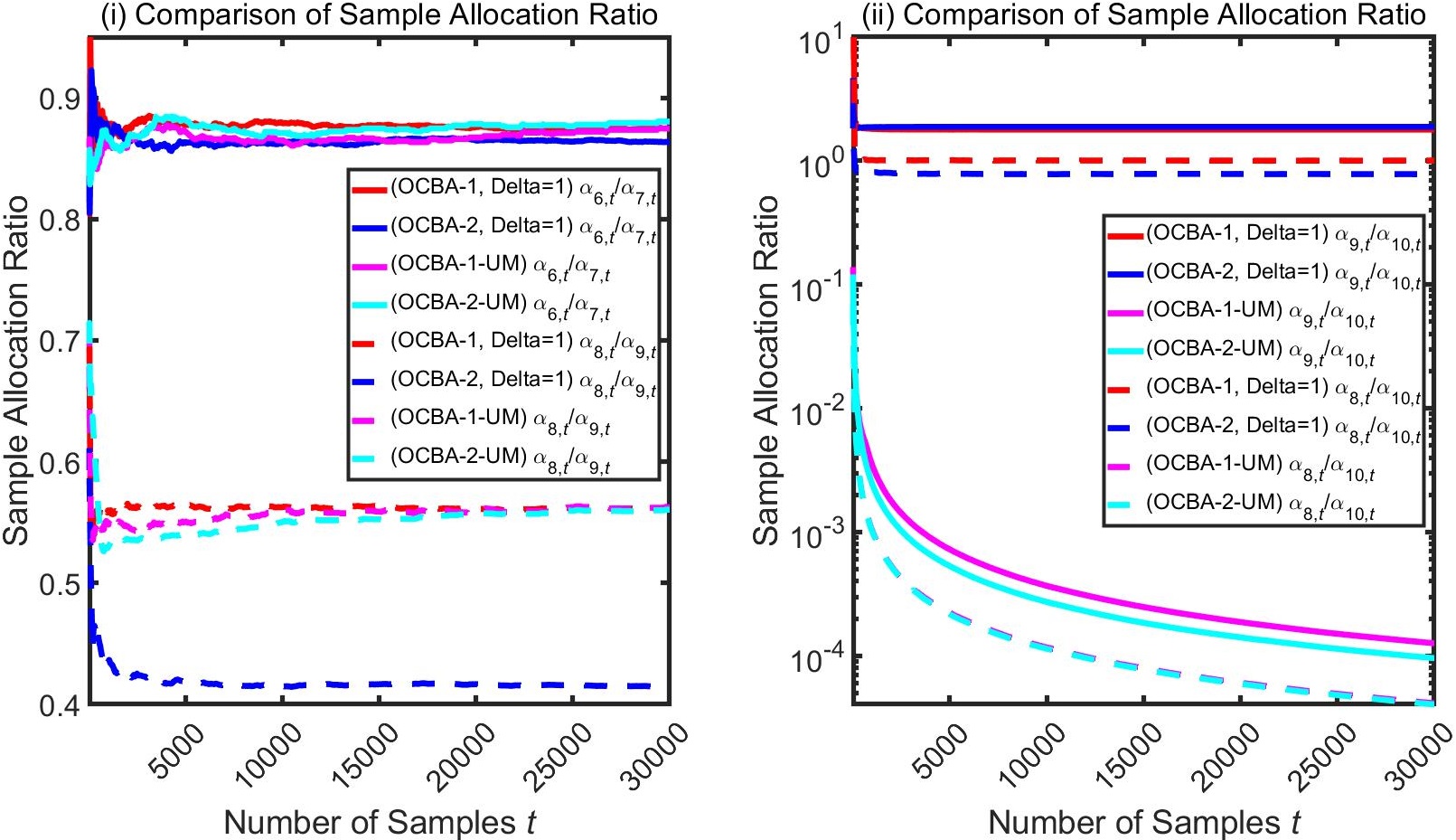}
	\end{subfigure}
	\begin{subfigure}
		\centering
		\includegraphics[scale=0.335]{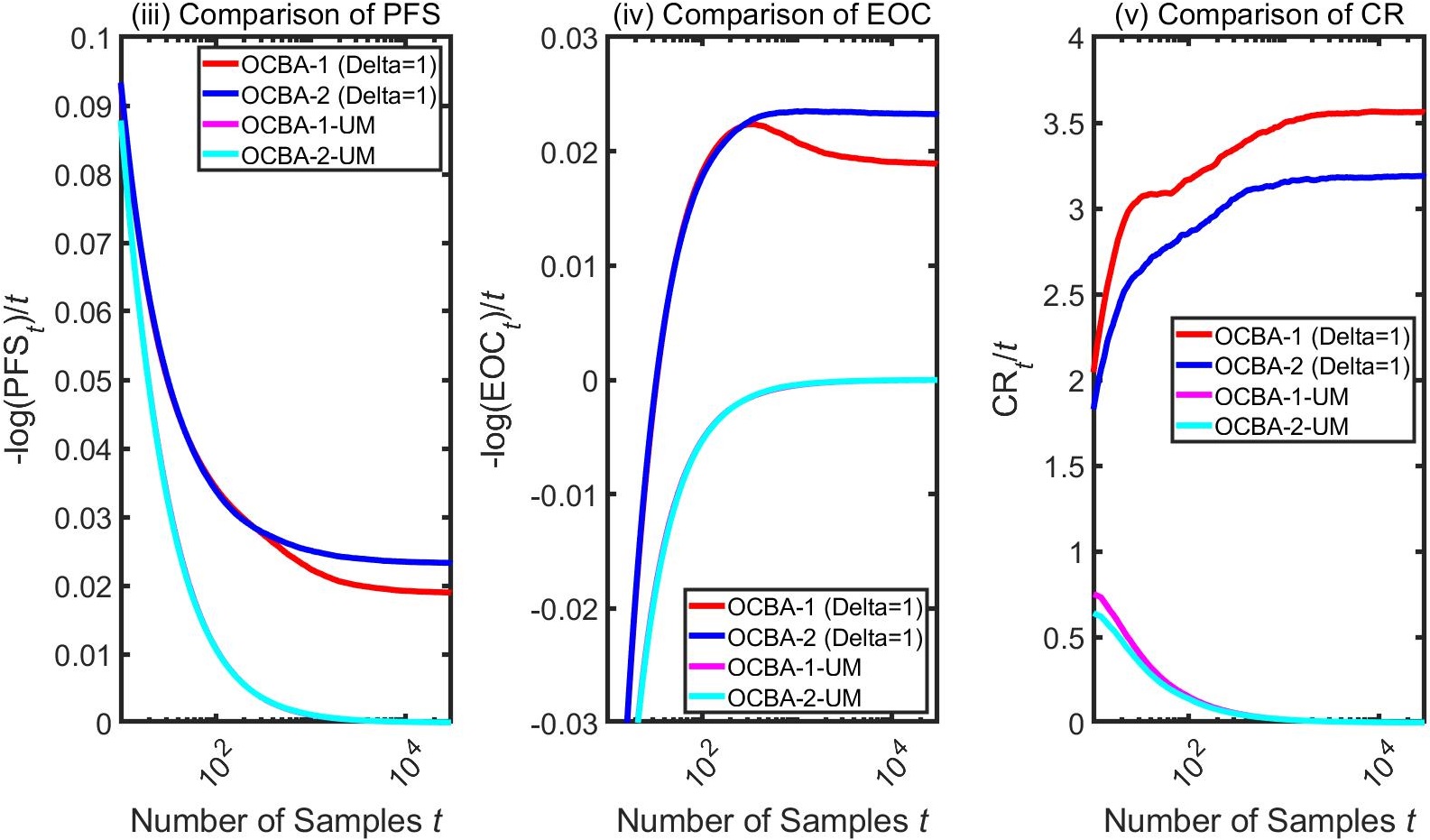}
	\end{subfigure}
	\caption{OCBA-1, OCBA-2, OCBA-1-UM, and OCBA-2-UM for instance 2.}
	\label{fig4}
\end{figure}

Figures \ref{fig5}(i) and \ref{fig5}(ii) show that the values of ${\rm PFS}_t$ and ${\rm EOC}_t$ for OCBA-1-UM and OCBA-2-UM are larger than those for Epsilon-Greedy and UCB1-Normal. OCBA-1-UM and OCBA-2-UM have almost the same values of ${\rm PFS}_t$ and ${\rm EOC}_t$. The similar pattern occurs for Epsilon-Greedy and UCB1-Normal. As the number of samples increases, the gaps among the four algorithms narrow in terms of ${\rm PFS}_t$ and ${\rm EOC}_t$. In Figure \ref{fig5}(iii), OCBA-1-UM and OCBA-2-UM have smaller values of ${\rm CR}_t$ than UCB1-Normal and Epsilon-Greedy. OCBA-1-UM and OCBA-2-UM have almost the same values of ${\rm CR}_t$. UCB1-Normal has the largest values of ${\rm CR}_t$.

\begin{figure}[!htbp]
	\vspace{0.5mm}
	\centering
	\includegraphics[scale=0.505]{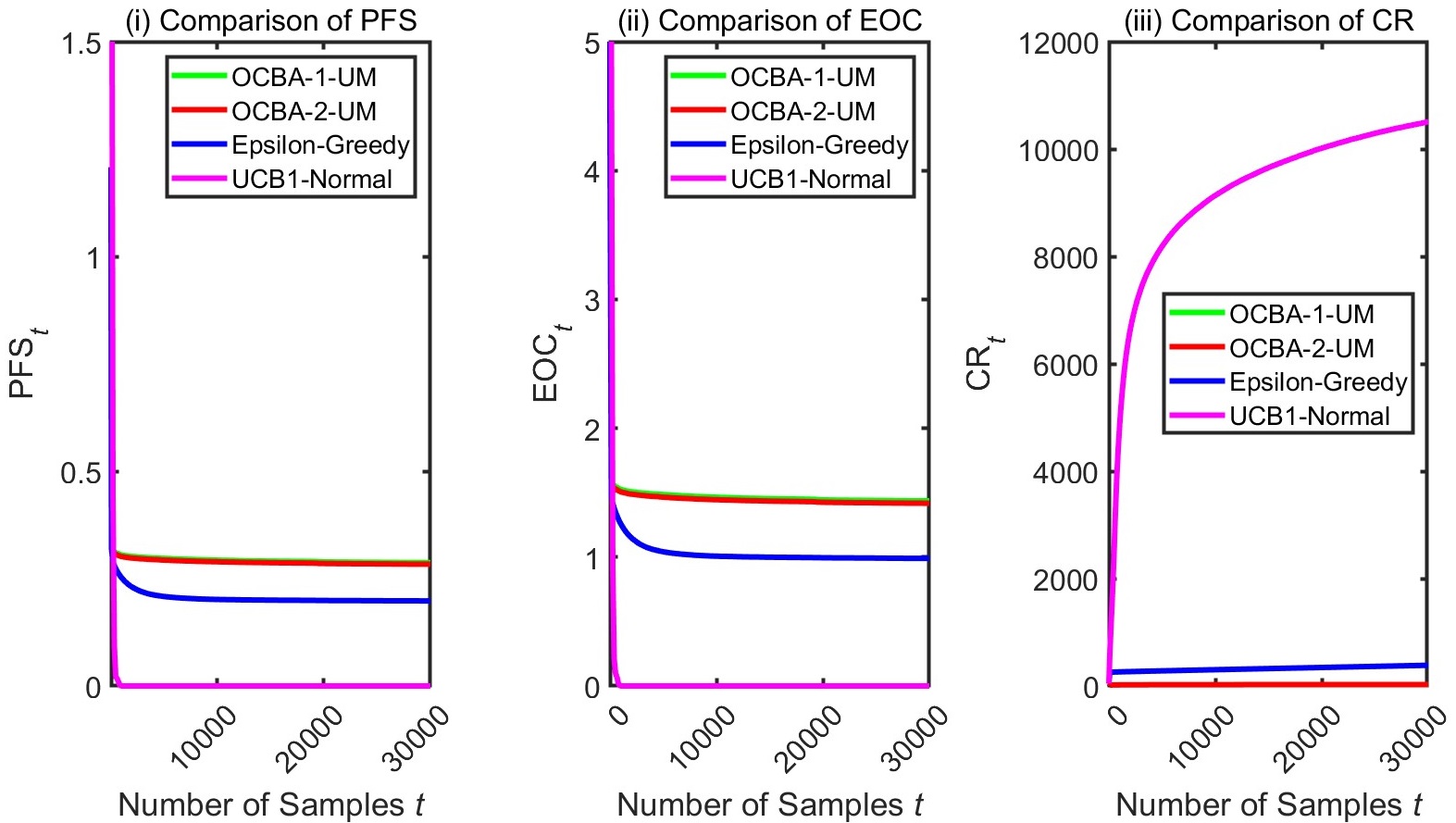}
	\caption{OCBA-1-UM, OCBA-2-UM, Epsilon-Greedy, and UCB1-Normal for instance 2.}
	\label{fig6}
\end{figure}

In Figures \ref{fig6}(i) and \ref{fig6}(ii), UCB1-Normal has the smallest values of ${\rm PFS}_t$ and ${\rm EOC}_t$ among the four algorithms. OCBA-1-UM and OCBA-2-UM have almost the same values of ${\rm PFS}_t$ and ${\rm EOC}_t$, and their ${\rm PFS}_t$ and ${\rm EOC}_t$ are larger than those of Epsilon-Greedy. Figure \ref{fig6}(iii) shows that the values of ${\rm CR}_t$ for UCB1-Normal are much larger than those of the other three algorithms. OCBA-1-UM and OCBA-2-UM have almost the same values of ${\rm CR}_t$, and their ${\rm CR}_t$ are smaller than that of Epsilon-Greedy.

We further discuss some patterns that appear in Figures \ref{fig1}-\ref{fig6}:

\begin{itemize}
	\item The performance of OCBA and OCBA-1 under PFS and EOC is competitive with or slightly better than that of OCBA-2 when there is a small number of samples. As the number of samples is large enough, the performance of OCBA-2 under PFS and EOC may surpass that of OCBA and OCBA-1. There is little difference between the performance of OCBA and OCBA-1 under PFS and EOC, no matter whether $\Delta=1$ or $\Delta=10$. The change of $\Delta$ has little effect on the performance of OCBA, OCBA-1, and OCBA-2 in terms of PFS and EOC. 
	
	\item The performance of OCBA-2 under CR is better than that of OCBA and OCBA-1, and OCBA performs the worst under CR among the three OCBA algorithms. In addition, the change from $\Delta=1$ to $\Delta=10$ results in worse performance of OCBA, OCBA-1 and OCBA-2 in terms of CR.
	
	\item OCBA-1, OCBA-1-UM, and OCBA-2-UM are almost the same in terms of the ratios of sample allocations for the non-best designs. It corroborates the theoretical findings of Theorem \ref{thm-ocbaum-sampalloc}(iii). For OCBA-1-UM and OCBA-2-UM, the sample allocation for the best design is much larger than that for each non-best one.
	
	\item If we evaluate the performance of the tested algorithms by PFS and EOC, OCBA-1-UM and OCBA-2-UM perform worse than OCBA-1, OCBA-2, Epsilon-Greedy, and UCB1-Normal. While in return, the performance of OCBA-1-UM and OCBA-2-UM under CR is superior to that of OCBA-1, OCBA-2, Epsilon-Greedy, and UCB1-Normal. That is, the improvement in the performance under CR is paralleled by the decline in the performance under PFS and EOC. This result aligns with the statement in \cite{bubeck2009} that the lower bound of PFS and EOC may be larger if the guaranteed upper bound of CR is smaller.
\end{itemize}

\section{Conclusions}\label{sec6}

In this paper, we study the OCBA-1 and OCBA-2 Algorithms under Gaussian samples and known variances of each design. We analyze the convergence rates of the two OCBA algorithms under three commonly used performance measures PFS, EOC and CR in the literature. It fills the gap of convergence analysis for algorithms designed based on the OCBA optimality conditions. The theoretical results confirm the high efficiency of the two algorithms under PFS and EOC that has been observed in numerical testing. Furthermore, we slightly modify the two OCBA algorithms and prove that the modified algorithms can achieve the optimal convergence rate under CR. It provides good insights into the relationship and difference between the measures PFS, EOC and CR and shows the potential of broader application of the OCBA-like algorithms.

In this research, we have assumed known variances for samples of each design. In practice, they are typically unknown and are replaced by sample variances. Then, an important future research direction is to analyze the performance of the OCBA algorithms under this setting. Wu and Zhou \cite{wu2018} did some exploratory work along this direction; however, some key questions still remain open, such as the asymptotic convergence rates and the upper and lower bounds of the algorithm performance under different measures. In addition, it may be possible to apply the framework of convergence analysis in this paper to analyze other OCBA algorithms, including the OCBA-m Algorithm \cite{chen2008} for subset selection, the OCBA-CO Algorithm \cite{lee2012} for selecting the best feasible design, etc. Another possible future direction is investigating how to relax the assumptions in developing the OCBA algorithms for broader application, such as relaxing the assumption of Gaussian samples to sub-Gaussian samples.

\bibliography{OCBA_Ref}
\bibliographystyle{plain}

\newpage
\appendix
\onecolumn

\renewcommand\thesection{A.\arabic{section}}
\setcounter{equation}{0}
\renewcommand\theequation{A.\arabic{equation}}

\section{Proof of Lemma \ref{lem1}}

\textit{Proof of (i).} Suppose that $\left\{\alpha_i^{(t)}\Big{|}i=1,\dots,k\right\}$ does not satisfy $\lim_{t\to\infty}\alpha_i^{\left(t\right)}=\alpha_i$ for $\forall i$. Then, $\exists i_0$, $\exists \epsilon_0$, $\forall T$, $\exists t_0>T$, 
\begin{align*}
\left|\alpha_{i_0}^{\left(t_0\right)}-\alpha_{i_0}\right|\geq\epsilon_0.
\end{align*}	
In addition, as $t\to\infty$, $N_i^{(t)}\to\infty$ for $\forall i$, and thus $M_i^{(t)}\to\infty$ for $\forall i$. Based on these conditions, we can find a subsequence $\left\{\alpha_{i}^{\left(t_p\right)}\Big{|}i=1,\dots,k\right\}$ satisfying that $M_i^{\left(p\right)}\to\infty$ as $p\to\infty$ for $\forall i$ and 
\begin{align*}
&\mathop{\lim\inf}_{p\to\infty}\alpha_{i_0}^{\left(t_p\right)}\geq\alpha_{i_0}+\epsilon_0\\
{\rm or}~ &\mathop{\lim\sup}_{p\to\infty}\alpha_{i_0}^{\left(t_p\right)}\leq\alpha_{i_0}-\epsilon_0.
\end{align*}
The subsequence $\left\{\alpha_i^{\left(t_p\right)}\Big{|}i=1,\dots,k\right\}$ does not have any subsequence whose convergence point is $\left\{\alpha_i\big{|}i=1,\dots,k\right\}$. It contradicts that each subsequence $\left\{\alpha_i^{\left(t_p\right)}\Big{|}M_i^{\left(p\right)}\to\infty\right.$ $\left.{\rm as}~p\to\infty,i=1,\dots,k\right\}$ has a further subsequence whose convergence point is $\left\{\alpha_i\big{|}i=1,\dots,k\right\}$. So $\lim_{t\to\infty}\alpha_i^{\left(t\right)}=\alpha_i$ for $\forall i$.
\\
\textit{Proof of (ii).} We prove it by contradiction. Without loss of generality, suppose that 
\begin{align*}
\underset{t\to\infty}{\lim}\left(\frac{\alpha_b^{\left(t\right)}}{\sigma_b}\right)^{2}-\underset{i\ne b}{\sum}\left(\frac{\alpha_i^{\left(t\right)}}{\sigma_i}\right)^{2}=0
\end{align*}
does not hold. It indicates that $\exists \epsilon_0$, $\forall T$, $\exists t_0>T$ such that 
\begin{align*}
\left|\left(\frac{\alpha_b^{\left(t_0\right)}}{\sigma_b}\right)^{2}-\underset{i\ne b}{\sum}\left(\frac{\alpha_i^{\left(t_0\right)}}{\sigma_i}\right)^{2}\right|\geq\epsilon_0.
\end{align*}
Following similar arguments as in the proof of Lemma \ref{lem1}(i), we can find a subsequence $\left\{\alpha_i^{\left(t_p\right)}\Big{|}i=1,\dots,k\right\}$ satisfying that $M_i^{\left(p\right)}\to\infty$ as $p\to\infty$ for $\forall i$ and 
\begin{align*}
&\underset{p\to\infty}{\lim\inf}\left(\frac{\alpha_b^{\left(t_p\right)}}{\sigma_b}\right)^{2}-\underset{i\ne b}{\sum}\left(\frac{\alpha_i^{\left(t_p\right)}}{\sigma_i}\right)^{2}\geq\epsilon_0\\
{\rm or}~ &\underset{p\to\infty}{\lim\sup}\left(\frac{\alpha_b^{\left(t_p\right)}}{\sigma_b}\right)^{2}-\underset{i\ne b}{\sum}\left(\frac{\alpha_i^{\left(t_p\right)}}{\sigma_i}\right)^{2}\leq-\epsilon_0.
\end{align*}
Then, $\left\{\alpha_i^{\left(t_p\right)}\Big{|}i=1,\dots,k\right\}$ does not have any subsequence that satisfies (\ref{lem1-eq1}). It contradicts that each subsequence $\left\{\alpha_i^{\left(t_p\right)}\Big{|}M_i^{\left(p\right)}\to\infty~{\rm as}~p\to\infty,i=1,\dots,k\right\}$ has a further subsequence satisfying (\ref{lem1-eq1}). So $\left\{\alpha_i^{(t)}\Big{|}i=1,\dots,k\right\}$ satisfies (\ref{lem1-eq1}).
\hfill $\square$

\section{Proof of Themrem \ref{thm-ocba1-sampalloc}}

We consider a probability space $\left(\Omega,\mathcal{F},\mathbb{P}\right)$ in which all the Gaussian random variables are well defined. $\exists\tilde{\Omega}\subseteq\Omega$, $\tilde{\Omega}$ is a measurable sample space and $\mathbb{P}\left(\tilde{\Omega}\right)=1$. According to the Strong Law of Large Numbers, $\forall\omega\in\tilde{\Omega}$, $\forall i\in\left\{1,\dots,k\right\}$, $\lim_{t\to\infty}\hat{\mu}_i^{(t)}=\mu_i$ if $N_i^{(t)}\to\infty$ as $t\to\infty$. For $\forall\omega\in\tilde{\Omega}$, $\hat{\mu}_i^{(t)}\ne\hat{\mu}_j^{(t)}$ occurs with probability one, $\forall i\ne j$, $i,j\in\left\{1,\dots,k\right\}$, because the Gaussian distributions are non-degenerate. Without loss of generality, we fix a sample path $\omega\in\tilde{\Omega}$. For notation simplicity, we omit the dependence of the terms on $\omega$ when there is no ambiguity. We also omit $n_0$ due to its little impact in asymptotic analysis.

\textit{Proof of (i).} Denote by $A=\left\{i\big{|}N_i^{(t)}\to\infty~{\rm as}~t\to\infty\right\}$, by $ B=\left\{1,\dots,k\right\}\setminus A$. $A\ne\emptyset$. To prove that $\lim_{t\to\infty}\hat{b}^{(t)}=b$, it is sufficient to prove that $B=\emptyset$. Suppose that $B\ne\emptyset$. For $\forall j\in B$, $\exists\xi_j>0$, design $j$ will not receive samples when $t>\xi_j$. Denote by $\hat{\alpha}_i\triangleq\lim_{t\to\infty}\frac{\hat{N}_i^{(t)}}{t}>0$ for $\forall i$. Then, $\sum_{i=1}^{k}\hat{\alpha}_i=1$, $\lim_{t\to\infty}\frac{\hat{N}_j^{(t)}-N_j^{(t)}}{t}=\hat{\alpha}_j$ for $\forall j\in B$. Let $\underline{\alpha}_i\triangleq\lim\inf_{t\to\infty}\frac{N_i^{(t)}}{t}$ for $\forall i\in A$, $\hat{\alpha}_{\rm Bmin}\triangleq\min_{j\in B}\hat{\alpha}_j$. Note that 
\begin{align}\label{thm1i-eq0}
\sum_{i\in A}\underline{\alpha}_i=1.
\end{align} 
Otherwise, $\sum_{i\in A}\underline{\alpha}_i<1$, and thus $\lim\sup_{t\to\infty}\frac{\sum_{j\in B}N_{j}^{(t)}}{t}$ $=1-\sum_{i\in A}\underline{\alpha}_i>0$. It means that $\exists j_0\in B$, $\exists T_0>\xi_{j_0}$, design $j_0$ will still receive samples when $t>T_0$, which contradicts $j_0\in B$. So (\ref{thm1i-eq0}) holds. In addition, we claim that for $\forall i\in A$, 
\begin{align}\label{thm1i-eq1}
\hat{\alpha}_i\geq\underline{\alpha}_i+\hat{\alpha}_{\rm Bmin}.
\end{align}
Otherwise, $\exists i_0\in A$ such that for $\forall j\in B$, $\underset{t\to\infty}{\lim\sup}\frac{\hat{N}_{i_0}^{(t)}-N_{i_0}^{(t)}}{t}$ $=\hat{\alpha}_{i_0}-\underline{\alpha}_{i_0}<\hat{\alpha}_{\rm Bmin}\leq\lim_{t\to\infty}\frac{\hat{N}_j^{(t)}-N_j^{(t)}}{t}$. It indicates that $\exists T_1>0$, design $i_0$ will not receive samples when $t>T_1$, which contradicts $i_0\in A$. So (\ref{thm1i-eq1}) holds. Combining (\ref{thm1i-eq0}), (\ref{thm1i-eq1}) and $\hat{\alpha}_{\rm Bmin}>0$, it holds that $\sum_{i\in A}\hat{\alpha}_i\geq\sum_{i\in A}\underline{\alpha}_i+\hat{\alpha}_{\rm Bmin}>1$, which contradicts $\sum_{i=1}^{k}\hat{\alpha}_i=1$. So $B=\emptyset$.

\textit{Proof of (ii).}
Suppose that $\left\{\alpha_i^{\left(t\right)}\Big{|}i=1,\dots,k\right\}$ does not converge. Based on Theorem \ref{thm-ocba1-sampalloc}(i), $M_i^{(t)}\to\infty$ as $t\to\infty$ for $\forall i$, where $M_i^{(t)}$ is from Lemma \ref{lem1}. We select any subsequence $\left\{\alpha_i^{\left(t_p\right)}\Big{|}M_i^{\left(p\right)}\to\infty~{\rm as}~p\to\infty,i=1,\dots,k\right\}$. According to Bolzano-Weierstrass theorem, we select a convergent subsequence,
denoted by $\left\{\alpha_i^{\left(t_{p_q}\right)}\Big{|}i=1,\dots,k\right\}$,
converging to $\left\{\alpha'_i\big{|}i=1,\dots,k\right\}$ and satisfying that $M_i^{\left(q\right)}\to\infty$ as $q\to\infty$ for $\forall i$. If $\left\{\alpha'_i\big{|}i=1,\dots,k\right\}\ne\left\{\alpha_i^{*}\big{|}i=1,\dots,k\right\}$, then $\exists i_0$, $\alpha_{i_0}^{*}<\alpha'_{i_0}$. So 
\begin{align*}
\hat{N}_{i_0}^{\left(t_{p_q}\right)}-N_{i_0}^{\left(t_{p_q}\right)}<0
\end{align*}
when $q$ is large enough. It indicates that $i_0$ receives at most finite samples in the limit, which contradicts that $M_{i_0}^{\left(q\right)}\to\infty$ as $q\to\infty$. That is, $\left\{\alpha'_i\big{|}i=1,\dots,k\right\}=\left\{\alpha_i^{*}\big{|}i=1,\dots,k\right\}$. Based on Lemma \ref{lem1}, $\lim_{t\to\infty}\alpha_i^{\left(t\right)}=\alpha_i^{*}$ for $\forall i$.
\hfill $\square$

\section{Proof of Theorem \ref{thm-ocba2-sampalloc}}

\textbf{Proof.} We fix a sample path $\omega\in\tilde{\Omega}$ and omit $\omega$, $n_0$ in the proof.

\textit{Proof of (i).} Denote by $A=\left\{i\big{|}N_i^{(t)}\to\infty~{\rm as}~t\to\infty\right\}$, by $ B=\left\{1,\dots,k\right\}\setminus A$. $A\ne\emptyset$. To prove that $\lim_{t\to\infty}\hat{b}^{(t)}=b$, it is sufficient to prove that $B=\emptyset$. Denote by $\hat{b}=\arg\max_{i=1,\dots,k}\left(\lim_{t\to\infty}\hat{\mu}_i^{(t)}\right)$. $\exists T_2$, $\forall t>T_2$, $\hat{b}^{(t)}=\hat{b}$. We first claim that $\hat{b}\in A$. Otherwise, $\exists \xi_{\hat{b}}$, $\hat{b}$ will not receive samples after $t>\xi_{\hat{b}}$. However, $A\ne\emptyset$ causes that $\lim_{t\to\infty}\left(\frac{N_{\hat{b}}^{\left(t\right)}}{\sigma_{\hat{b}}}\right)^{2}-\sum_{j\ne\hat{b}}\left(\frac{N_j^{\left(t\right)}}{\sigma_j}\right)^{2}<0$. That is, $\hat{b}$ will still be sampled after $t>\xi_{\hat{b}}$. It contradicts that $\hat{b}\in B$. So $\hat{b}\in A$. We can similarly prove that $A\setminus\left\{\hat{b}\right\}\ne\emptyset$. We claim that $B=\emptyset$. Otherwise, for $i\in B$, $j\in A$, $j\ne \hat{b}$, 
\begin{align*}
\frac{\left(\hat{\mu}_i^{\left(t\right)}-\hat{\mu}_{\hat{b}}^{\left(t\right)}\right)^{2}}{\frac{\sigma_i^{2}}{N_i^{\left(t\right)}}+\frac{\sigma_{\hat{b}}^{2}}{N_{\hat{b}}^{\left(t\right)}}}<\infty,~
\frac{\left(\hat{\mu}_j^{\left(t\right)}-\hat{\mu}_{\hat{b}}^{\left(t\right)}\right)^{2}}{\frac{\sigma_j^{2}}{N_j^{\left(t\right)}}+\frac{\sigma_{\hat{b}}^{2}}{N_{\hat{b}}^{\left(t\right)}}}\to \infty
\end{align*}
as $t\to\infty$. It contradicts that $i\in B$. So $B=\emptyset$.

\textit{Proof of (ii).} Based on Theorem \ref{thm-ocba2-sampalloc}(i), $M_i^{(t)}\to\infty$ as $t\to\infty$ for $\forall i$. Without loss of generality, we select any subsequence $\left\{\alpha_i^{\left(t_p\right)}\bigg{|}i=1,\dots,k\right\}$ that satisfies $M_i^{\left(p\right)}\to\infty$ as $p\to\infty$ for $\forall i$. According to Bolzano-Weierstrass theorem, we select a convergent subsequence, denoted by $\left\{\alpha_i^{\left(t_{p_q}\right)}\Big{|}i=1,\dots,k\right\}$, converging to $\left\{\alpha'_i\big{|}i=1,\dots,k\right\}$ and satisfying that $M_i^{\left(q\right)}\to\infty$ as $q\to\infty$. For the OCBA-2 Algorithm, $\left(\frac{\alpha'_b}{\sigma_b}\right)^{2}-\sum_{i\ne b}\left(\frac{\alpha'_i}{\sigma_i}\right)^{2}<0$ contradicts that $M_i^{\left(q\right)}\to\infty$ as $q\to\infty$. In the same way, $\left(\frac{\alpha'_b}{\sigma_b}\right)^{2}-\sum_{i\ne b}\left(\frac{\alpha'_i}{\sigma_i}\right)^{2}>0$ does not hold. So 
\begin{align*}
\left(\frac{\alpha'_b}{\sigma_b}\right)^{2}-\sum_{i\ne b}\left(\frac{\alpha'_i}{\sigma_i}\right)^{2}=0.
\end{align*}
Based on this condition, $\alpha'_b>0$ and $\alpha'_i>0$ for some $i\ne b$. We claim that $\alpha'_i>0$ for all $i$. Otherwise, for $i\ne b$ satisfying $\alpha'_i=0$ and $j\ne b$ satisfying $\alpha'_j>0$, $\left(\frac{\sigma_j^{2}}{N_j^{\left(t_{p_q}\right)}}+\frac{\sigma_b^{2}}{N_b^{\left(t_{p_q}\right)}}\right)\bigg{/}\left(\frac{\sigma_i^{2}}{N_i^{\left(t_{p_q}\right)}}+\frac{\sigma_b^{2}}{N_b^{\left(t_{p_q}\right)}}\right)\to 0$ as $q\to\infty$. It indicates that $j$ will receive at most finite samples in the limit. It contradicts that $\alpha'_j>0$. So $\alpha'_i>0$ for $\forall i$. Furthermore, we claim that 
\begin{align*}
\frac{\left(\mu_b-\mu_i\right)^{2}}{\frac{\sigma_i^{2}}{\alpha'_i}+\frac{\sigma_b^{2}}{\alpha'_b}}=\frac{\left(\mu_b-\mu_j\right)^{2}}{\frac{\sigma_j^{2}}{\alpha'_j}+\frac{\sigma_b^{2}}{\alpha'_b}},
\end{align*}
$\forall i,j\ne b$. Without loss of generality, suppose that $\exists i_0,j_0\ne b$, $i_0\ne j_0$, 
\begin{align*}
\frac{\left(\mu_b-\mu_{i_0}\right)^{2}}{\frac{\sigma_{i_0}^{2}}{\alpha'_{i_0}}+\frac{\sigma_b^{2}}{\alpha'_b}}>\frac{\left(\mu_b-\mu_{j_0}\right)^{2}}{\frac{\sigma_{j_0}^{2}}{\alpha'_{j_0}}+\frac{\sigma_b^{2}}{\alpha'_b}}.
\end{align*}
It implies that $i_0$ will receive at most finite samples in the limit. It contradicts that $M_{i_0}^{\left(q\right)}\to\infty$ as $q\to\infty$. So the claim holds. That is, $\left\{\alpha_i^{(t)}\big{|}i=1,\dots,k\right\}$ satisfies the conditions of Lemma \ref{lem1}. We know that the strictly concave optimization problem (\ref{opt-pfs}) has a global optimum solution which is unique, denoted by $\left\{\alpha_i^{**}\big{|}i=1,\dots,k\right\}$. According to \cite{boyd2004} and \cite{glynn2004}, the KKT conditions are necessary and sufficient for the global optimization of (\ref{opt-pfs}). In addition, an alternative form of the KKT conditions is (\ref{ocba-glynn}) and $\sum_{i=1}^{k}\alpha_i=1$. So $\left\{\alpha_i^{**}\big{|}i=1,\dots,k\right\}$ is the unique solution of (\ref{ocba-glynn}) and $\sum_{i=1}^{k}\alpha_i=1$. Based on this condition, we can apply the same arguments as used above and prove by contradiction that $\lim_{t\to\infty}\alpha_i^{(t)}=\alpha_i^{**}$ for $\forall i$.
\hfill $\square$

\section{Proof of Theorem \ref{thm-ocba1-convrate}}

We fix a sample path $\omega\in\tilde{\Omega}$ and omit $\omega$, $n_0$ in the proof.

\textit{Proof of (i).} For each design $i$, denote by $\tilde{\mu}_i^{(t)}$ the random variable whose distribution is the distribution of the sample mean of design $i$ in the $t$-th iteration when given no specific sample value. Based on Bonferroni inequality \cite{galambos1977}, 
\begin{align*}
{\rm PFS}_t=\mathbb{P}\left(\bigcup_{i\ne b}\left(\tilde{\mu}_b^{(t)}<\tilde{\mu}_i^{(t)}\right)\right)\leq\sum_{i\ne b}\mathbb{P}\left(\tilde{\mu}_b^{(t)}<\tilde{\mu}_i^{(t)}\right).
\end{align*}
It then holds that 
\begin{equation}\label{thm3-neq1}
\begin{aligned}
&\max_{i\ne b}\mathbb{P}\left(\tilde{\mu}_b^{(t)}<\tilde{\mu}_i^{(t)}\right)\\
\leq&{\rm PFS}_t\leq\left(k-1\right)\max_{i\ne b}\mathbb{P}\left(\tilde{\mu}_b^{(t)}<\tilde{\mu}_i^{(t)}\right).
\end{aligned}
\end{equation}
Based on \cite{chen2011}, for $\forall i\ne b$,
\begin{align}\label{thm3-eq1}
\mathbb{P}\left(\tilde{\mu}_b^{(t)}<\tilde{\mu}_i^{(t)}\right)=\Phi\left(-\sqrt{\eta_i^{(t)}t}\right),
\end{align} 
where $\Phi(\cdot)$ is the cumulative density function of standard Gaussian distribution, $\eta_i^{(t)}=\frac{\left(\mu_b-\mu_i\right)^{2}\Delta}{\frac{\sigma_i^{2}}{\alpha_i^{(t)}}+\frac{\sigma_b^{2}}{\alpha_b^{(t)}}}$. According to \cite{gordon1941}, $\forall x>0$,
\begin{align}\label{thm3-neq2}
\frac{x}{\sqrt{2\pi}\left(1+x^{2}\right)}\exp\left\{-\frac{x^{2}}{2}\right\}\leq\Phi(x)\leq\frac{1}{\sqrt{2\pi}x}\exp\left\{-\frac{x^{2}}{2}\right\}.
\end{align}
Based on (\ref{thm3-neq1}), (\ref{thm3-eq1}) and (\ref{thm3-neq2}),
\begin{align*}
&\min_{i\ne b}\left(\frac{\sqrt{\frac{\eta_i^{(t)}t}{2\pi}}}{1+\eta_i^{(t)}t}\right)\max_{i\ne b}\left(\exp\left\{-\frac{\eta_i^{(t)}t}{2}\right\}\right)\\
\leq&{\rm PFS}_t\leq\frac{k-1}{\min_{i\ne b}\left(\sqrt{2\pi\eta_i^{(t)}t}\right)}\max_{i\ne b}\left(\exp\left\{-\frac{\eta_i^{(t)}t}{2}\right\}\right).
\end{align*}
Then, we can derive that $\lim_{t\to\infty}\frac{\log\left({\rm PFS}_t\right)}{-\min_{i\ne b}\left(\frac{\eta_i^{(t)}t}{2}\right)}=1$. It means that 
\begin{align*}
{\rm PFS}_t\doteq\exp\left\{-\min_{i\ne b}\left(\frac{\eta_i^{(t)}t}{2}\right)\right\}.
\end{align*}
In addition, 
\begin{align*}
\exp\left\{-\min_{i\ne b}\left(\frac{\eta_i^{(t)}t}{2}\right)\right\}\doteq\exp\left\{-\frac{\eta^{*}t}{2}\right\},
\end{align*}
where $\eta^{*}=\underset{i\ne b}{\min}\frac{\left(\mu_b-\mu_i\right)^{2}\Delta}{\frac{\sigma_i^{2}}{\alpha_i^{*}}+\frac{\sigma_b^{2}}{\alpha_b^{*}}}$, because $\lim_{t\to\infty}\alpha_i^{(t)}=\alpha_i^{*}$ for $\forall i$. According to transitivity of logarithmic equivalence, 
\begin{align*}
{\rm PFS}_t\doteq\exp\left(-\frac{\eta^{*}t}{2}\right).
\end{align*}

\textit{Proof of (ii).} Notice that $\left(\mu_b-\max_{i\ne b}\mu_i\right){\rm PFS}_t
\leq{\rm EOC}_t
\leq\left(\mu_b-\min_{i\ne b}\mu_i\right){\rm PFS}_t$. So ${\rm EOC}_t=\Theta\left({\rm PFS}_t\right)$, which implies that ${\rm EOC}_t\doteq{\rm PFS}_t$. Based on ${\rm PFS}_t\doteq\exp\left(-\frac{\eta^{*}t}{2}\right)$, it holds that 
\begin{align*}
{\rm EOC}_t\doteq\exp\left(-\frac{\eta^{*}t}{2}\right).
\end{align*}

\textit{Proof of (iii).} According to \cite{bubeck2012}, ${\rm CR}_t=\sum_{i\ne b}\left(\mu_b-\mu_i\right)\mathbb{E}\left[N_i^{(t)}\right]$. It then holds that
\begin{align*}
\underset{t\to\infty}{\lim}\frac{{\rm CR}_t}{\sum_{i\ne b}\left(\mu_b-\mu_i\right)\alpha_i^{*}\Delta t}
=\underset{t\to\infty}{\lim}\frac{\sum_{i\ne b}\left(\mu_b-\mu_i\right)\mathbb{E}\left[\frac{N_i^{(t)}}{\Delta t}\right]}{\sum_{i\ne b}\left(\mu_b-\mu_i\right)\alpha_i^{*}}.
\end{align*}
Based on Theorem \ref{thm-ocba1-sampalloc},  $0<\frac{N_i^{(t)}}{\Delta\cdot t}<1$, $\lim_{t\to\infty}\frac{N_i^{(t)}}{\Delta\cdot t}=\alpha_i^{*}$ for $\forall i$. According to Lebesgue's dominated convergence theorem, $\lim_{t\to\infty}\mathbb{E}\left[\frac{N_i^{(t)}}{\Delta\cdot t}\right]=\mathbb{E}\left[\lim_{t\to\infty}\frac{N_i^{(t)}}{\Delta\cdot t}\right]=\alpha_i^{*}$. Thus, ${\rm CR}_t\cong\sum_{i\ne b}\left(\mu_b-\mu_i\right)\alpha_i^{*}\Delta\cdot t$.
\hfill $\square$

\section{Proof of Theorem \ref{thm-ocba2-convrate}}

We fix a sample path $\omega\in\tilde{\Omega}$ and omit $\omega$, $n_0$ in the proof.

\textit{Proof of (i).} By similar discussion in the proof of Theorem \ref{thm-ocba1-convrate}(i),
\begin{align*}
{\rm PFS}_t\doteq\exp\left\{-\min_{i\ne b}\left(\frac{\eta_i^{(t)}t}{2}\right)\right\}.
\end{align*}
Based on Theorem \ref{thm-ocba2-sampalloc}, $\lim_{t\to\infty}\alpha_i^{(t)}=\alpha_i^{**}$ for $\forall i$. Then, $\exp\left\{-\min_{i\ne b}\left(\frac{\eta_i^{(t)}t}{2}\right)\right\}\doteq\exp\left\{-\frac{\eta^{**}t}{2}\right\}$, where $\eta^{**}=\min_{i\ne b}\left(\frac{\left(\mu_b-\mu_i\right)^{2}\Delta}{\frac{\sigma_i^{2}}{\alpha_i^{**}}+\frac{\sigma_b^{2}}{\alpha_b^{**}}}\right)$. Thus, 
\begin{align*}
{\rm PFS}_t\doteq\exp\left(-\frac{\eta^{**}t}{2}\right).
\end{align*}

\textit{Proof of (ii).} Following similar arguments as in the proof of Theorem \ref{thm-ocba1-convrate}(ii), ${\rm PFS}_t\doteq{\rm EOC}_t$. In addition, ${\rm PFS}_t\doteq\exp\left(-\frac{\eta^{**}t}{2}\right)$. So 
\begin{align*}
{\rm EOC}_t\doteq\exp\left(-\frac{\eta^{**}t}{2}\right).
\end{align*}

\textit{Proof of (iii).} By similar discussion in the proof of Theorem \ref{thm-ocba1-convrate}(iii), 
\begin{align*}
{\rm CR}_t=\sum_{i\ne b}\left(\mu_b-\mu_i\right)\mathbb{E}\left[N_i^{(t)}\right],~ \underset{t\to\infty}{\lim}\frac{\mathbb{E}\left[N_i^{(t)}\right]}{\Delta\cdot t}=\alpha_i^{**}.
\end{align*}
So ${\rm CR}_t\cong\sum_{i\ne b}\left(\mu_b-\mu_i\right)\alpha_i^{**}\Delta\cdot t$.
\hfill$\square$

\section{Proof of Theorem \ref{thm-ocbaum-sampalloc}}

\textbf{Proof (OCBA-1-UM).} In the $t$-th iteration, the OCBA-1-UM Algorithm provides one additional sample. The sampling strategy comprises two parts:
\begin{itemize}
	\item[] \textbf{Part 1:} With probability $\epsilon_t=\min\left\{\frac{h_t}{t},1\right\}$, it allocates the sample to one of the designs using the OCBA-1 allocation rule (line 6 of the OCBA-1-UM Algorithm).
	
	\item[] \textbf{Part 2:} With probability $1-\epsilon_t$, it allocates the sample to the estimated best design (line 8 of the OCBA-1-UM Algorithm).
\end{itemize}
Denote by $\tilde{N}_{\text{Part1}}^{(t)}$ the total number of samples allocated based on Part 1 of the OCBA-1-UM Algorithm until the $t$-th iteration, by $\tilde{N}_{\text{Part2}}^{(t)}$ the total number of samples allocated based on Part 2 of the OCBA-1-UM Algorithm until the $t$-th iteration. $\exists\overline{h}>\underline{h}>0$, $\underline{h}\log\left(\frac{t+1}{\underline{h}}\right)\leq\mathbb{E}\left[\tilde{N}_{\text{Part1}}^{(t)}\right]\leq\overline{h}\left(1+\log t\right)$, $\mathbb{E}\left[\tilde{N}_{\text{Part2}}^{(t)}\right]\geq t-\overline{h}\left(1+\log t\right)$ for large enough $t$. Denote by $\mathcal{E}_s$ the event when $\tilde{N}_{\text{Part1}}^{(s)}=\tilde{N}_{\text{Part1}}^{(s-1)}+1$. $\tilde{N}_{\text{Part1}}^{(0)}=0$. Let $\mathbbm{1}\left[\mathcal{E}_s\right]$ be the indicator function of $\mathcal{E}_s$. Denote by $V_s=\mathbb{P}\left(\mathcal{E}_s\right)-\mathbbm{1}\left[\mathcal{E}_s\right]$,
\begin{align*}
S_t=\sum_{s=1}^{t}V_s,~K_t=\sum_{s=1}^{t}{\rm Var}\left[\mathbbm{1}\left[\mathcal{E}_s\right]\right]\leq\mathbb{E}\left[\tilde{N}_{\text{Part1}}^{(t)}\right],
\end{align*} 
Based on Lemma \ref{lem3},
\begin{align*}
&\mathbb{P}\left(\tilde{N}_{\text{Part1}}^{(t)}\leq\left(1-\frac{1}{\log\left(\log t\right)}\right)\mathbb{E}\left[\tilde{N}_{\text{Part1}}^{(t)}\right]\right)\\
=&\mathbb{P}\left(S_t\geq\frac{\mathbb{E}\left[\tilde{N}_{\text{Part1}}^{(t)}\right]}{\log\left(\log t\right)},K_t\leq\mathbb{E}\left[\tilde{N}_{\text{Part1}}^{(t)}\right]\right)\\
\leq&\exp\left\{-\frac{\mathbb{E}\left[\tilde{N}_{\text{Part1}}^{(t)}\right]}{2\left(\log\left(\log t\right)\right)^{2}+\frac{2}{3}\log\left(\log t\right)}\right\}.
\end{align*} 
Then, $\mathbb{P}\left(\underset{t\to\infty}{\lim\inf}\frac{\tilde{N}_{\text{Part1}}^{(t)}}{\left(1-\frac{1}{\log\left(\log t\right)}\right)\mathbb{E}\left[\tilde{N}_{\text{Part1}}^{(t)}\right]}\leq 1\right)=0$ because $\mathbb{E}\left[\tilde{N}_{\text{Part1}}^{(t)}\right]\geq \underline{h}\log\left(\frac{t+1}{\underline{h}}\right)$ as $t\to\infty$. Furthermore, let $V'_t=-V_t$ and $S'_t=-S_t$. We can similarly prove that $\mathbb{P}\left(\underset{t\to\infty}{\lim\sup}\frac{\tilde{N}_{\text{Part1}}^{(t)}}{\left(1+\frac{1}{\log\left(\log t\right)}\right)\mathbb{E}\left[\tilde{N}_{\text{Part1}}^{(t)}\right]}\geq 1\right)=0$. Thus, $\underset{t\to\infty}{\lim}\frac{\tilde{N}_{\text{Part1}}^{(t)}}{\mathbb{E}\left[\tilde{N}_{\text{Part1}}^{(t)}\right]}\overset{a.s.}{=}1$. We can similarly prove that $\underset{t\to\infty}{\lim}\frac{\tilde{N}_{\text{Part2}}^{(t)}}{\mathbb{E}\left[\tilde{N}_{\text{Part2}}^{(t)}\right]}\overset{a.s.}{=}1$. We consider any fixed sample path $\omega\in\tilde{\Omega}$ that satisfies $\underset{t\to\infty}{\lim}\frac{\tilde{N}_{\text{Part1}}^{(t)}}{\mathbb{E}\left[\tilde{N}_{\text{Part1}}^{(t)}\right]}=\underset{t\to\infty}{\lim}\frac{\tilde{N}_{\text{Part2}}^{(t)}}{\mathbb{E}\left[\tilde{N}_{\text{Part2}}^{(t)}\right]}=1$. We omit $\omega$ and $n_0$ in the proof.

\textit{Proof of (i).} Denote by $A=\left\{i\big{|}N_i^{(t)}\to\infty~{\rm as}~t\to\infty\right\}$, by $ B=\left\{1,\dots,k\right\}\setminus A$. $A\ne\emptyset$. To prove that $\lim_{t\to\infty}\hat{b}^{(t)}=b$, it is sufficient to prove that $B=\emptyset$. Denote by $\hat{b}=\arg\max_{i=1,\dots,k}\left(\lim_{t\to\infty}\hat{\mu}_i^{(t)}\right)$. $\exists T_2$, $\forall t>T_2$, $\hat{b}^{(t)}=\hat{b}$. We claim that $\hat{b}\in A$. Otherwise, $\exists\xi_{\hat{b}}$, $\hat{b}$ will not receive samples after $t>\xi_{\hat{b}}$. However, $\tilde{N}_{\text{Part2}}^{(t)}\to\infty$ as $t\to\infty$. That is, $\hat{b}$ will still receive samples based on Part 2 of the OCBA-1-UM Algorithm after $t>\xi_{\hat{b}}$. It contradicts that $\hat{b}\in B$. So $\hat{b}\in A$. Next, we claim that $A\setminus\left\{\hat{b}\right\}\ne\emptyset$. Otherwise, for $\forall i\ne\hat{b}$, $\exists\xi_{i}$, $i$ will not be sampled after $t>\xi_{i}$. Meanwhile, $\lim_{t\to\infty}\frac{N_{\hat{b}}^{(t)}}{t}=1$ causes that $\hat{b}$ will only receive finite samples based on Part 1 of the OCBA-1-UM Algorithm. However, $\tilde{N}_{\text{Part1}}^{(t)}\to\infty$ as $t\to\infty$. It contradicts that $i\in B$ for $\forall i\ne\hat{b}$. So $A\setminus\left\{\hat{b}\right\}\ne\emptyset$. Lastly, we can prove that $B=\emptyset$ by similar discussion in the proof of Theorem \ref{thm-ocba1-sampalloc}(i).

\textit{Proof of (ii).} We know that $\lim_{t\to\infty}\hat{b}^{(t)}=b$. It indicates that each non-best design will receive at most finite samples based on Part 2 of the OCBA-1-UM Algorithm. So 
\begin{align*}
\lim_{t\to\infty}\frac{N_b^{(t)}}{t}\geq 1-\lim_{t\to\infty}\frac{\tilde{N}_{\text{Part1}}^{(t)}}{t}=1.
\end{align*}

\textit{Proof of (iii).} For $\forall i\ne b$, $M_i^{(t)}\to\infty$ as $t\to\infty$. Denote by $\tilde{\alpha}_i^{(t)}=\frac{N_i^{(t)}}{\sum_{j\ne b}N_j^{(t)}}$ for $\forall i\ne b$. We claim that
\begin{align*}
\lim_{t\to\infty}\tilde{\alpha}_i^{(t)}=\frac{\alpha_i^{*}}{\sum_{j\ne b}\alpha_j^{*}},~\forall i\ne b.
\end{align*}
By similar discussion in the proof of Theorem \ref{thm-ocba1-sampalloc}(ii), any subsequence $\left\{\tilde{\alpha}_i^{\left(t_p\right)}\bigg{|}M_i^{\left(p\right)}\to\infty~{\rm as}~p\to\infty,i\ne b\right\}$ has a subsequence that converges to $\left\{\frac{\alpha_i^{*}}{\sum_{j\ne b}\alpha_j^{*}}\bigg{|}i\ne b\right\}$. According to Lemma \ref{lem1}, $\lim_{t\to\infty}\tilde{\alpha}_i^{(t)}=\frac{\alpha_i^{*}}{\sum_{j\ne b}\alpha_j^{*}}$ holds for $\forall i\ne b$. Then, $\mathbb{E}\left[\tilde{N}_{\text{Part1}}^{(t)}\right]\cong h^{*}\left(\sum_{i\ne b}\alpha_i^{*}\right)\log t$. Notice that $b$ will receive at most finite samples based on Part 1 of the OCBA-1-UM Algorithm, and non-best designs will receive at most finite samples based on Part 2 of the OCBA-1-UM Algorithm. Based on this condition, we can derive that $\lim_{t\to\infty}\frac{N_i^{(t)}}{h^{*}\log t}=\alpha_i^{*}$, $\forall i\ne b$.

\textbf{Proof (OCBA-2-UM).} In the $t$-th iteration, the OCBA-2-UM Algorithm provides one additional sample. The sampling strategy comprises two parts: 
\begin{itemize}
	\item[] \textbf{Part 1:} With probability $\epsilon_t=\min\left\{\frac{h_t}{t},1\right\}$, it allocates the sample to one of the designs using the OCBA-2 allocation rule (line 6 of the OCBA-2-UM Algorithm).
	
	\item[] \textbf{Part 2:} With probability $1-\epsilon_t$, it allocates the sample to the estimated best design (line 8 of the OCBA-2-UM Algorithm).
\end{itemize}
Denote by $\tilde{N}_{\text{Part1'}}^{(t)}$ the total number of samples allocated based on Part 1 of the OCBA-2-UM Algorithm until the $t$-th iteration, by $\tilde{N}_{\text{Part2'}}^{(t)}$ the total number of samples allocated based on Part 2 of the OCBA-2-UM Algorithm until the $t$-th iteration. By similar discussion as used above, $\underset{t\to\infty}{\lim}\frac{\tilde{N}_{\text{Part1'}}^{(t)}}{\mathbb{E}\left[\tilde{N}_{\text{Part1'}}^{(t)}\right]}\overset{a.s.}{=}1$, $\underset{t\to\infty}{\lim}\frac{\tilde{N}_{\text{Part2'}}^{(t)}}{\mathbb{E}\left[\tilde{N}_{\text{Part2'}}^{(t)}\right]}\overset{a.s.}{=}1$. We fix any sample path $\omega\in\tilde{\Omega}$ satisfying 
\begin{align*}
\underset{t\to\infty}{\lim}\frac{\tilde{N}_{\text{Part1'}}^{(t)}}{\mathbb{E}\left[\tilde{N}_{\text{Part1'}}^{(t)}\right]}=\underset{t\to\infty}{\lim}\frac{\tilde{N}_{\text{Part2'}}^{(t)}}{\mathbb{E}\left[\tilde{N}_{\text{Part2'}}^{(t)}\right]}=1.
\end{align*}
We omit $\omega$ and $n_0$ in the proof.

\textit{Proof of (i).} It can be shown by similar discussion of Theorems \ref{thm-ocba2-sampalloc}(i) and \ref{thm-ocbaum-sampalloc}(i) for the OCBA-1-UM Algorithm. 

\textit{Proof of (ii).} Notice that $\lim_{t\to\infty}\frac{\tilde{N}_{\text{Part1'}}^{(t)}}{t}=0$. We also know that each non-best design will receive at most finite samples based on Part 2 of the OCBA-2-UM Algorithm. So $\lim_{t\to\infty}\frac{N_b^{(t)}}{t}\geq 1-\lim_{t\to\infty}\frac{\tilde{N}_{\text{Part1'}}^{(t)}}{t}=1$.

\textit{Proof of (iii).} For $\forall i$, as $t\to\infty$, $N_i^{(t)}\to\infty$ leads to $M_i^{(t)}\to\infty$. Denote by $\tilde{\alpha}_i^{(t)}=\frac{N_i^{(t)}}{\sum_{j\ne b}N_j^{(t)}}$ for $\forall i\ne b$. We claim that  $\lim_{t\to\infty}\tilde{\alpha}_i^{(t)}=\frac{\alpha_i^{*}}{\sum_{j\ne b}\alpha_j^{*}}$ for $\forall i\ne b$. We select any subsequence $\left\{\tilde{\alpha}_i^{\left(t_p\right)}\Big{|}M_i^{\left(p\right)}\to\infty~{\rm as}~p\to\infty,i\ne b\right\}$ of $\left\{\tilde{\alpha}_i^{(t)}\Big{|}i\ne b\right\}$. Without loss of generality, we select a convergence point of it, denoted by $\left\{\alpha'_i\big{|}i\ne b\right\}$. If $\left\{\alpha'_i\big{|}i\ne b\right\}\ne\left\{\frac{\alpha_i^{*}}{\sum_{j\ne b}\alpha_j^{*}}\bigg{|}i\ne b\right\}$, then $\exists i_0,j_0\ne b$, $i_0\ne j_0$, 
\begin{align*}
\frac{\left(\mu_b-\mu_{i_0}\right)^{2}}{\sigma_{i_0}^{2}}\alpha'_{i_0}>\frac{\left(\mu_b-\mu_{j_0}\right)^{2}}{\sigma_{j_0}^{2}}\alpha'_{j_0}.
\end{align*}
Notice that $\lim_{t\to\infty}\frac{N_i^{(t)}}{N_b^{(t)}}=0$ for $\forall i\ne b$. 
It implies that
\begin{align*}
\frac{\left(\hat{\mu}_b^{\left(t_{p_q}\right)}-\hat{\mu}_{i_0}^{\left(t_{p_q}\right)}\right)^{2}}{\frac{\sigma_{i_0}^{2}}{N_{i_0}^{\left(t_{p_q}\right)}}+\frac{\sigma_{b}^{2}}{N_{b}^{\left(t_{p_q}\right)}}}>\frac{\left(\hat{\mu}_b^{\left(t_{p_q}\right)}-\hat{\mu}_{j_0}^{\left(t_{p_q}\right)}\right)^{2}}{\frac{\sigma_{j_0}^{2}}{N_{j_0}^{\left(t_{p_q}\right)}}+\frac{\sigma_{b}^{2}}{N_{b}^{\left(t_{p_q}\right)}}}
\end{align*}
for large enough $t_{p_q}$. It reaches a contradiction that $M_{i_0}^{\left(q\right)}\to\infty$ as $q\to\infty$. So $\left\{\alpha'_i\big{|}i\ne b\right\}=\left\{\frac{\alpha_i^{*}}{\sum_{j\ne b}\alpha_j^{*}}\bigg{|}i\ne b\right\}$. Based on Lemma \ref{lem1}, $\lim_{t\to\infty}\tilde{\alpha}_i^{(t)}=\frac{\alpha_i^{*}}{\sum_{j\ne b}\alpha_j^{*}}$ for $\forall i\ne b$. Then, we can similarly show that $\lim_{t\to\infty}\frac{N_i^{(t)}}{h^{*}\log t}=\alpha_i^{*}$ for $\forall i\ne b$.
\hfill $\square$

\section{Proof of Theorem \ref{thm-ocbaum-convrate}}

Here we only show the proof for the OCBA-1-UM Algorithm. Statements for the OCBA-2-UM Algorithm can be similarly proved.

Consider any fixed sample path $\omega\in\tilde{\Omega}$ that satisfies $\underset{t\to\infty}{\lim}\frac{\tilde{N}_{\text{Part1}}^{(t)}}{\mathbb{E}\left[\tilde{N}_{\text{Part1}}^{(t)}\right]}=1$ and $\underset{t\to\infty}{\lim}\frac{\tilde{N}_{\text{Part2}}^{(t)}}{\mathbb{E}\left[\tilde{N}_{\text{Part2}}^{(t)}\right]}=1$. We omit $\omega$ and $n_0$ in the proof.

\textit{Proof of (i).} Based on $\underset{t\to\infty}{\lim}\frac{N_i^{(t)}}{N_b^{(t)}}=0$ for $\forall i\ne b$ and similar arguments as in the proof of Theorem \ref{thm-ocba1-convrate}(i), 
\begin{align*}
{\rm PFS}_t\doteq\exp\left\{-\frac{1}{2}\min_{i\ne b}\left(\frac{\left(\mu_b-\mu_i\right)^{2}}{\sigma_i^{2}}N_i^{(t)}\right)\right\}.
\end{align*}
Based on Theorem \ref{thm-ocbaum-sampalloc}(iii), ${\rm PFS}_t\doteq t^{-\frac{h^{*}}{2}\min_{i\ne b}\left(\frac{\left(\mu_b-\mu_i\right)^{2}}{\sigma_i^{2}}\alpha_i^{*}\right)}$, which can be re-expressed as ${\rm PFS}_t\doteq t^{-\frac{h^{*}}{2\sum_{i=1}^{k}\beta_i}}$.

\textit{Proof of (ii).} By similar discussion in the proof of Theorem \ref{thm-ocba1-convrate}(ii),  ${\rm PFS}_t\doteq{\rm EOC}_t$. In addition, ${\rm PFS}_t\doteq t^{-\frac{h^{*}}{2\sum_{i=1}^{k}\beta_i}}$. So ${\rm EOC}_t\doteq t^{-\frac{h^{*}}{2\sum_{i=1}^{k}\beta_i}}$.

\textit{Proof of (iii).} Notice that
\begin{align*}
\frac{{\rm CR}_t}{h^{*}\sum_{i\ne b}\left(\mu_b-\mu_i\right)\alpha_i^{*}\log t}=\frac{\sum_{i\ne b}\left(\mu_b-\mu_i\right)\frac{\mathbb{E}\left[N_i^{(t)}\right]}{h^{*}\log t}}{\sum_{i\ne b}\left(\mu_b-\mu_i\right)\alpha_i^{*}}.
\end{align*}
By similar discussion in the proof of Theorem \ref{thm-ocba1-convrate}(iii), $\lim_{t\to\infty}\frac{\mathbb{E}\left[N_i^{(t)}\right]}{h^{*}\log t}=\alpha_i^{*}$. So ${\rm CR}_t\cong
\left(\sum_{i\ne b}\frac{\mu_b-\mu_i}{{\rm kl}_{i,b}}\right)\log t$.
\hfill $\square$

\end{document}